\documentclass[journal]{IEEEtran} %a4paper

\usepackage{cite}

\ifCLASSINFOpdf
  \usepackage[pdftex]{graphicx}
  \DeclareGraphicsExtensions{.pdf,.jpeg,.png}
\else
  \usepackage[dvips]{graphicx}
  \DeclareGraphicsExtensions{.eps}
\fi
\usepackage{algorithm}
\usepackage{algpseudocode}
\usepackage[hidelinks]{hyperref} % added to make tilde work in URLs
\usepackage{dblfloatfix}
\usepackage{multirow}

\usepackage[table]{xcolor}
\usepackage{colortbl}
\usepackage{amsmath,amssymb,amsthm,mathrsfs,amsfonts,dsfont} 
\usepackage{bm}
\usepackage{makecell}
\usepackage{rotating}
\usepackage[switch]{lineno}
\begin{document}

\def \mytitle {Memetic Viability Evolution\\for Constrained Optimization}

\def \algorithmname {mViE}

\title{\mytitle}

\author{Andrea~Maesani, Giovanni~Iacca,~\IEEEmembership{Member,~IEEE}, and~Dario~Floreano,~\IEEEmembership{Senior~Member,~IEEE}% <-this % stops a space
\thanks{A. Maesani, G. Iacca and D. Floreano are with the Laboratory of Intelligent Systems, 
Institute of Microengineering, \'{E}cole Polytechnique F\'{e}d\'{e}rale de Lausanne (EPFL), Lausanne, 1015, Switzerland
e-mail: \{andrea.maesani, giovanni.iacca, dario.floreano\}@epfl.ch}% <-this % stops a space
\thanks{Manuscript received August 28, 2014; revised December 29, 2014 and March 11, 2015; accepted April 25, 2015.}
\thanks{Copyright (c) 2015 IEEE. Personal use of this material is permitted. However, permission to use this material for any other purposes must be obtained from the IEEE by sending a request to \href{mailto:pubs-permissions@ieee.org}{pubs-permissions@ieee.org}.}}


\markboth{IEEE Transactions on Evolutionary Computation,~Vol.~XX, No.~XX, August~XXXX}%
{Maesani \MakeLowercase{\textit{et al.}}: \mytitle}

\IEEEpubid{0000--0000/00\$00.00~\copyright~2015 IEEE}


\maketitle

\begin{abstract}
The performance of evolutionary algorithms can be heavily undermined when constraints limit the feasible areas of the search space. For instance, while Covariance Matrix Adaptation Evolution Strategy is one of the most efficient algorithms for unconstrained optimization problems, it cannot be readily applied to constrained ones. Here, we used concepts from Memetic Computing, i.e. the harmonious combination of multiple units of algorithmic information, and Viability Evolution, an alternative abstraction of artificial evolution, to devise a novel approach for solving optimization problems with inequality constraints. Viability Evolution emphasizes elimination of solutions not satisfying viability criteria, defined as boundaries on objectives and constraints. These boundaries are adapted during the search to drive a population of local search units, based on Covariance Matrix Adaptation Evolution Strategy, towards feasible regions. These units can be recombined by means of Differential Evolution operators. Of crucial importance for the performance of our method, an adaptive scheduler toggles between exploitation and exploration by selecting to advance one of the local search units and/or recombine them. The proposed algorithm can outperform several state-of-the-art methods on a diverse set of benchmark and engineering problems, both for quality of solutions and computational resources needed.
\end{abstract}

\begin{IEEEkeywords}
Constrained Optimization, Covariance Matrix Adaptation, Differential Evolution, Memetic Computing, Viability Evolution.
\end{IEEEkeywords}

% \fi
\IEEEpeerreviewmaketitle

\section{Introduction}\label{sec:intro}
\IEEEPARstart{S}{everal} real-world optimization problems are characterized by the presence of one or more inequality constraints that limit the feasible region of the search space. 
When defined in a continuous domain, such problems can be formulated as:
\begin{equation}
\min{f(\bm{x})}, \textrm{  s.t. :} 
\begin{cases}
\begin{array}{ll}
l_{i} \leq x_{i} \leq u_{i}, & i=1,2,\ldots,n\\
g_{j}(\bm{x})\leq 0, & j=1,2,\ldots,m
\end{array}
\end{cases}
\end{equation}
where $f(\bm{x})$ is the objective (fitness) function to be optimized and $\bm{x}\in\mathbb{R}^n$ is a vector of 
design variables $[x_{1},x_{2},\ldots,x_{n}]$. The search space is delimited by box constraints defining the admissible range $[l_{i},u_{i}], l_{i} \in \mathbb{R}, u_{i} \in \mathbb{R}$ of each variable $x_{i}, i \in 1,2,\ldots,n$. Within the search space, the inequality constraints $g_{j}(\bm{x})$ determine the feasible region that contains all the solutions satisfying the constraints of the problem.\footnote{Constrained optimization problems can also include equality constraints defined as $h_{k}(\bm{x})=0, k=1,2,\ldots,p$, that can reduce the feasible areas down to zero-volume regions. Dedicated literature covers this class of problems proposing specific techniques for handling them. Handling equality constraints is out of the scope of this paper.} 

Examples of constrained optimization problems can be found in many fields where physical, geometrical or resource requirements may limit the feasibility of the solutions. Due to the vast range of applications, constrained optimization has attracted the interest of a large part of the Computational Intelligence research community\footnote{A constantly updated list of references on the topic, maintained by Carlos A. Coello Coello, is available at: \url{http://www.cs.cinvestav.mx/~constraint/}.}.
Pioneering studies can be traced back to research on Evolutionary Algorithms (EA), particularly Genetic Algorithms (GA) \cite{Michalewicz96b,Rasheed98anadaptive,Coello2000,DBLP:journals/aei/CoelloM02} and Evolution Strategies (ES) \cite{bib:mezura2005}, while more recently Particle Swarm Optimization (PSO) has gained attention \cite{bib:hu2003,Aguirre:2007,He:2007:ECP:1219187.1219657,Cagnina08a,Pant:2009:LDI:1670662.1670669,DBLP:journals/eswa/Coelho10}, likewise other Swarm Intelligence techniques \cite{Leguizamon09,bib:yang2010,Brajevic_SC_ABC,Cuevas2014412,DBLP:journals/tcyb/ZhangYH14}.
Parallel research lines have explored the use of several constraint handling techniques \cite{Michalewicz95asurvey,coello2002theoretical,Mezura11}, as adaptive penalty functions \cite{Tessema09,Kramer2013}, repair mechanisms for infeasible solutions \cite{Salcedo09,Wessing2013}, stochastic ranking of solutions \cite{Runarsson2000}, $\varepsilon$-constrained optimization \cite{Takahama10}, feasibility rules to rank solutions \cite{Deb2000}, and surrogate models \cite{jin:surrogate-assisted}. Recent empirical studies analyzed the utility of retaining infeasible solutions during evolution \cite{conf/cec/WhileH13,WangCai2008}, whereas others have considered the use of multi-objective techniques where the constraint violations are minimized as separate objectives together with the problem objective function \cite{Angantyr,constrainedMOEA,Clevenger2005,CaiWang2006}. Importantly, competitions for constrained optimization organized in the context of the IEEE Congress on Evolutionary Computation (CEC) \cite{ProbCEC2006,ProbCEC2010} have finally provided standard benchmarks for comparing the performance of the various algorithms in the field.

Despite these advances, the increasing number of computationally intensive applications still need to be matched by computationally efficient algorithms. This is of utmost importance in contexts where evaluations are computationally expensive, as in the case of complex simulations, or where the optimization process has to be run in a limited time, as in the case of hardware-in-the-loop evolution. In these scenarios, optimization algorithms should ideally deliver high performance (in terms of quality of discovered solutions) using only a limited amount of function evaluations. \IEEEpubidadjcol
However, state-of-the-art Computational Intelligence algorithms for constrained optimization typically require a large number of evaluations to converge, especially if the problem is high-dimensional, multi-modal and severely constrained.

Among the most effective algorithms for single objective optimization, Covariance Matrix Adaptation Evolution Strategy (CMA-ES) \cite{Hansen-CMAES} has gained considerable attention in the last decade due to its ability of solving highly non-separable, ill-conditioned, and multi-modal functions. Although some attempts of extending CMA-ES to constrained optimization problems have been made \cite{bib:Kramer2009,Collange2010,Kusakci2013a,Gieseke2013,deMelo2014}, CMA-ES is not yet competitive on these types of problems: the self-adaptation of the algorithm's parameters is not suitable in constrained landscapes \cite{Beyer2012}. Moreover, due to the use of a single search distribution in CMA-ES, it is difficult to explore disconnected feasible areas unless restarts occur.

Arnold and Hansen \cite{Arnold2012} recently introduced in a (1+1)-CMA-ES \cite{Igel2006} a novel covariance matrix adaptation rule. This rule exploits constraint violations to learn information and adapt the covariance matrix to decrease the likelihood of sampling other infeasible solutions. However, the method works only when an initial feasible solution is provided. In \cite{viePPSN14} we extended this method to allow its use also when started from infeasible solutions, by taking inspiration from Viability Evolution principles \cite{Mattiussi2003,maesani2014artificial}. Viability Evolution is an alternative abstraction of artificial evolution that operates by eliminating individuals not satisfying a set of criteria. These criteria, called viability boundaries, are defined on the problem objectives or constraints, and are adapted during evolution. Similar to what is done in a Viability Evolution algorithm, in \cite{viePPSN14} the viability boundaries defined on the constraints of the problem are relaxed or tightened to drive the search towards feasible areas. This method, called (1+1)-ViE-CMA-ES, or in short (1+1)-ViE, was also enriched with a novel mechanism to adapt the step-size based on information collected at each constraint violation. The method displayed excellent performance on unimodal constrained optimization problems but, since it is based on a version of CMA-ES that samples a single offspring per generation, its performance was quite poor on problems characterized by disconnected feasible areas or highly multi-modal fitness landscapes. 

%even if this would surely result in a method more robust to multi-modality, it 
One possible way to partially overcome these limitations is to reformulate the rules for covariance matrix \cite{Arnold2012} and step-size adaptation \cite{viePPSN14} designed for (1+1)-CMA-ES into rules for $(\mu,\lambda)$-CMA-ES. However, this solution would still rely on a single search distribution, thus having limited capability of exploring disconnected feasible areas. An alternative solution is the use of multiple independent local search units made of (1+1)-ViE, that may potentially explore disconnected feasible regions. The main burden of this second option is represented by the need for an efficient algorithm that determines the optimal allocation of function evaluations among the local search units, as simply running all of them in parallel may result in a waste of computational resources. On the other hand, once an appropriate algorithmic scheme for balancing the function evaluations among the different local search units is found, it could be possible to re-use it also with different types of local searchers.

In this paper, we explored this latter idea by proposing a novel memetic computing approach, called \textit{memetic Viability Evolution} (\algorithmname), that uses several local search units, constituted by (1+1)-ViE \cite{viePPSN14}, that can be recombined by Differential Evolution (DE) \cite{jour:storn1997} evolutionary operators. Differential Evolution was chosen as global search operator due to its 
intrinsic ability of adjusting the effect of mutations during the search. Such feature makes DE a good candidate for recombining the information learned by the local search units, while performing at the same time global exploration. This intuition was also supported by previous research \cite{bib:mezura2007} that showed through a large comparative study that DE tends to outperform several alternative meta-heuristics (such as PSO, GA, and ES) on constrained optimization problems. Furthermore, we improved the balance between exploration and exploitation by using an adaptive scheduler that allocates dynamically the function evaluations to either the local search units or to global search.

To assess \algorithmname, we performed numerical experiments on thirteen CEC 2006 benchmark problems with inequality constraints \cite{ProbCEC2006}, as well as four classical mechanical engineering design problems\cite{Coello2000,Gandomi11}. We then compared our method against an extensive collection of state-of-the-art algorithms for constrained optimization. In the experiments, \algorithmname~showed consistent performance gains in terms of function evaluations needed to reach the optimum on almost all the tested problems.

The paper is organized as follows. Section \ref{sec:related} surveys the literature related to constrained optimization. Section \ref{sec:related-oneplusone} briefly describes (1+1)-CMA-ES and recent advances for the adaptation of covariance matrix and step-size. Moreover, this section discusses the introduction of Viability Evolution principles in CMA-ES. Section~\ref{sec:algorithm} describes in details the proposed method. The experimental setup and the algorithmic parameter setting are discussed in section~\ref{sec:setup}, while the numerical results are presented in section~\ref{sec:results}. Finally, discussion and conclusions are presented in section~\ref{sec:conclusion}.
\section{Related Work: Constrained Optimization}\label{sec:related}

In this section we first review related literature on constrained optimization algorithms based on CMA-ES and DE, the building blocks of our proposed method \algorithmname. Then, we summarize recent advances in memetic computing, focusing on approaches specifically designed for constrained problems. 

\subsection{Methods based on CMA-ES}
Covariance Matrix Adaptation Evolution Strategy (CMA-ES)~\cite{Hansen-CMAES} is considered nowadays the state-of-the-art in unconstrained single-objective optimization. In the presence of constraints, however, the step-size control used by CMA-ES to refine the search does not work properly \cite{Beyer2012}, an effect known also in standard ES. Recent research efforts have been devoted to overcome this difficulty. The use of adaptive penalty functions was investigated in \cite{Collange2010}, where the weight of each constraint in the penalty function was modified according to the number of iterations during which that constraint was violated. Another penalty function has been proposed in \cite{deMelo2014}, where the constraint violation of all the solutions in the population is used to scale the relative violation of each solution. CMA-ES has been also integrated with ASCHEA, an approach proposed in \cite{Hamida02} to adapt the tolerances on the equality constraints \cite{Kusakci2013a}. Other approaches rank individuals based on three independent rankings \cite{Kusakci2013}, namely objective function, constraint violation, and number of violated constraints, or use surrogate models to learn information about constraints \cite{bib:Kramer2009, Gieseke2013}. A repair mechanism was used in a problem-specific variant of CMA-ES for financial optimization \cite{Beyer2012}. 

Among the most effective approaches to date for CMA-ES-based constrained optimization, Arnold and Hansen \cite{Arnold2012} proposed a modification of (1+1)-CMA-ES \cite{Igel2006} specifically designed for unimodal problems. Starting from a feasible solution, the algorithm maintains a low-pass filtered vector representing the direction of violations of each constraint and consequently uses this information to reduce the variance of the search ellipsoid along the detected direction of violation. In our previous study \cite{viePPSN14}, we further improved the performance of this (1+1)-CMA-ES scheme, by collecting information on single constraint violations and using it to adapt the step-size. 

\subsection{Methods based on Differential Evolution}
Earlier research on DE-based constrained optimization considered the use of feasibility rules and diversity preservation mechanisms \cite{bib:mezura2004,Mezura2006} and the incorporation of domain knowledge \cite{LandaBecerra20064303}. With the introduction of the CEC benchmarks on constrained optimization \cite{ProbCEC2006,ProbCEC2010}, that created a common environment for assessing the performance of novel constrained optimization algorithms, DE became a popular choice for solving constrained problems. A number of DE-based methods proposed for solving these benchmarks are now considered the state-of-the-art in evolutionary constrained optimization. Among these methods, $\varepsilon$-DE \cite{Takahama:COb:cec2006} ranks, by objective value, the solutions that are feasible or violate at most by an $\varepsilon$-value the constraints, while preferring feasible solutions over infeasible solutions, which are instead compared based on their constraint violation. Notably, $\varepsilon$-DE won both the CEC 2006 and 2010 competitions for constrained optimization. Competitive results were also obtained by two other variants of DE, namely MDE \cite{Mezura-Montes:MDE:cec2006}, which uses an \emph{ad-hoc} mutation that incorporates information from both best and parent individuals, and SADE \cite{Huang:SDE:cec2006}, a self-adaptive version of DE. Both algorithms employ the three feasibility rules presented in \cite{Deb2000} for handling constraints. 
%In a comparative study \cite{bib:mezura2007}, DE was shown to outperform PSO, GA, and ES also on various numerical and engineering problems. %MOVED TO THE INTRO
Later research on optimal DE parameter control \cite{bib:mezura2009a,Mezura10} led to devise robust self-adapting schemes, such as those presented in \cite{bib:mezura2009b} and \cite{Zou11a}. Furthermore, specific mutation operators were introduced in \cite{Mohamed:2012:COB:2181343.2181778/COMDE,journals/soco/KongOP13}. Also, DE was combined with (adaptive) penalty functions in \cite{bib:Melo2012,RePEc:spr:coopap:v:54:y:2013:i:3:p:707-739}, Lagrangian methods for handling equality constraints in \cite{journals/cad/LongLHC13}, or even ensembles of constraint handling techniques \cite{Mallipeddi10}. More recently, $\varepsilon$-DE was further extended through the introduction of ranking \cite{DBLP:conf/cec/TakahamaS12a} and surrogate models by using kernel regression \cite{conf/cec/TakahamaS13}.

\subsection{Memetic Computing approaches}
Lastly, it is worth mentioning recent research on constrained optimization by Memetic Computing (MC) approaches. Memetic Computing is an emerging trend in Computational Intelligence whose focus is on algorithms composed of multiple interacting operators, also named \emph{memes}. Originally inspired by the diffusion of ideas in evolving populations of learning agents, where individuals undergo genetic evolution but also show learning capabilities and adaptation \cite{moscato1989evolution}, MC nowadays embraces a broad plethora of methods consisting of multiple (possibly heterogeneous) units of ``algorithmic information'' whose synergistic coordination is used for problem-solving \cite{ong2010frontier}. 
% ORIGINAL QUOTE: `` units of information encoded in computational representations for the purpose of problem-solving''
As such, the modern MC concept is not limited only to optimization problems. It also goes far beyond the initial definition of Memetic Algorithms (MA), where a population-based algorithm (e.g. an EA) is coupled with one or more individual learning units (local search) \cite{moscato1989evolution}, as it includes several algorithmic schemes characterized by various levels of adaptation and decision-making \cite{ong2004metalamlearn,ong2006,nguyen2009}, coevolution \cite{smith2007}, local surrogate models \cite{zhoy2007}, or purposely simple cascades of single-solution search units \cite{iacca2012ockham}. For a thorough survey of MC, we refer the interested reader to \cite{chen2011,bib:Neri2012Handbook}. 
%other stuff by Ong: \cite{ong2009SI} (special issue on MA), \cite{Aydt2011} (UML for MC), \cite{Chen2012} (meme complexes)

So far, MC has been successfully applied mostly to unconstrained optimization, either continuous or combinatorial \cite{bib:Neri2012Handbook}. On the other hand, only few studies tackled constrained problems by means of memetic techniques. Presumably, this might be due to the difficulty in coordinating and balancing global exploration with local search, a crucial aspect in MC \cite{ishibuchi2003balance}, when the fitness landscape is highly constrained and the search approaches the boundary of the feasible region. An interesting example of a MC method for constrained optimization is given by the agent-based memetic algorithm \cite{Ullah09b,Ullah011}, where a society of agents co-evolves by recombination only and each individual selects, independently, a local search technique from a predefined pool. 
%\cite{Ullah2007,Ullah08,Ullah09a,Ullah12}
Similar ideas can be found in \cite{Pescador2012}, where global search (by means of DE operator), stochastic ranking and a crossover-based local search are coordinated in a memetic fashion. Other hybrid algorithms combine GA and Artificial Immune System (AIS) \cite{Bernardino07}, PSO and Simulated Annealing (SA) \cite{He07a}, PSO and GA \cite{Takahama05b}, PSO and DE \cite{Liu:2010:HPS:1660170.1660493}. In two $(\mu$+$\lambda$)-DE approaches \cite{Jia2013302,Wang11a}, multiple mutation operators are applied while the algorithms modify at run-time the policy used for ranking solutions, according to the composition of the population in terms of feasible/infeasible solutions.

More recently, some studies proposed extremely powerful memetic algorithms that make use of gradient-based information, either on the fitness or on the constraints, thus making an implicit assumption of continuous and differentiable functions. An example of such methods is given in \cite{SunGaribaldi2010}, where an Estimation of Distribution Algorithm (EDA) is combined with a classic gradient-based local optimizer (DONLP2). Similarly, in \cite{Handoko2010} a memetic algorithm is proposed that combines a GA, Sequential Quadratic Programming (SQP) with second-order functional approximations, and modelling of the feasibility region through Support Vector Machine. A consensus-based variant of GA is combined with SQP also in \cite{doi:10.1080/0305215X.2013.846336}, but in this case the algorithm uses gradient information on the constraints. Finally, a hybrid algorithm combining PSO, DE, CMA-ES, gradient-based mutation and constraint handling with $\varepsilon$-level comparison was proposed in \cite{Bonyadi:2013:HPS:2463372.2463378}.

\section{Related Work: (1+1)-ViE-CMA-ES}\label{sec:related-oneplusone}

In this section we summarize the main concepts behind covariance matrix \cite{Arnold2012} and step-size \cite{viePPSN14} adaptation in CMA-ES applied to constrained optimization. Furthermore, we summarize our previous work on the application of Viability Evolution principles to (1+1)-CMA-ES \cite{viePPSN14}, resulting in the method named (1+1)-ViE. 

\subsection{(1+1)-CMA-ES for constrained optimization}
Arnold and Hansen \cite{Arnold2012} introduced a variant of (1+1)-CMA-ES for constrained optimization. In the (1+1)-CMA-ES, an offspring solution $\bm{y} \leftarrow \bm{x} + \sigma \bm{A}\bm{z}$ is sampled at each generation from the parent solution $\bm{x} \in \mathbb{R}^n$, where $\bm{z} \sim \mathcal{N}(0,\mathbf{I})$, $\bm{A}$ is the Choleski decomposition of the covariance matrix $\bm{C} = \bm{A}^T \bm{A}$, and $\sigma \in \mathbb{R}_+$ is the global step-size. 
During the search, the step-size $\sigma$ and the covariance matrix $\bm{C}$ are adapted using information learned while sampling solutions. %in the search space.

The step-size $\sigma$ is adapted with the 1/5\textsuperscript{th} rule \cite{Igel2006}, which increases or decreases $\sigma$ observing the probability of generating successful solutions $P_{succ} \in [0,1]$, as follows:
\begin{equation}
\label{ref:step-size-adaptation}
\sigma \leftarrow \sigma \cdot exp\left( \frac{P_{succ} - \frac{P_{target}}{1 - P_{target}}(1 - P_{succ}) }{d}\right)
\end{equation}
where the damping factor $d$ controls the step-size variation, and $P_{target}$ marks the threshold between decrease or increase in $\sigma$.
At each iteration, $P_{succ}$ is updated as follows:
\begin{equation}
P_{succ} \leftarrow (1-c_p)P_{succ} + c_p \mathds{1}_{f(\bm{y}) \leq f(\bm{x})}
\end{equation}
where $\mathds{1}_{f(\bm{y}) \leq f(\bm{x})}$ is 1 if the offspring solution improves or equals the parent solution or 0 otherwise, and the learning rate $c_p \in (0,1]$ determines the fading of $P_{succ}$. 
	
The covariance matrix is adapted using two separate rules. The first rule, the original rank-one update of CMA-ES, is applied whenever an offspring solution improves the parent. In this case, the variance in the direction $\bm{v}$ is increased from one iteration $g$ to the following one as in: % reported in Equation \ref{eq:original-rank-one}
\begin{equation}
\label{eq:original-rank-one}
\bm{C}^{(g+1)} = \alpha \bm{C}^{(g)} + \beta \bm{v}^{(g)}\bm{v}^{(g)^T}
\end{equation}

The (1+1)-CMA-ES uses as direction $\bm{v}$ the vector of fading successful steps $\bm{s}$, called evolution path, which is updated based on the value of the probability $P_{succ}$, and the parameters $c$, and $c_{cov}^+$ defined in Table \ref{tab:oneplusoneparameters}. More specifically, if $P_{succ} < P_{thresh}$ then $\bm{s} \gets (1-c)\bm{s} + \sqrt{c(2-c)}\bm{A}\bm{z}$ and $\alpha = 1-c_{cov}^+$. On the contrary, if $P_{succ} \geq P_{thresh}$ then $\bm{s} \gets (1-c)\bm{s}$ and $\alpha = 1-c_{cov}^+ + c_{cov}^+c(2-c)$. In both cases, $\beta$ is set equal to $c_{cov}^+$. This update increases the likelihood of sampling offspring solutions in the direction of previous successful steps. Rather than updating the covariance matrix using the original rank-one update (Equation \ref{eq:original-rank-one}), the Choleski factor $\bm{A}$ can be directly modified as shown in \cite{Igel2006}:
\begin{equation}\label{eq:cholupdate}
\bm{A} \leftarrow \sqrt{\alpha} \bm{A} + \frac{\sqrt{\alpha}}{\|\bm{w}\|^2}\left( \sqrt{1 + \frac{\beta}{\alpha}\|\bm{w}\|^2} - 1 \right) \bm{s}\bm{w}^T
\end{equation}
where $\bm{w} = \bm{A}^{-1}\bm{s}$. %\in \mathbb{R}^n$
In addition, the algorithm of Arnold and Hansen \cite{Arnold2012} uses also the ``active" covariance matrix update rule presented in \cite{Arnold2010}. The covariance matrix is updated to decrease the variance in the direction of a particularly unsuccessful step, specifically whenever an offspring is worse than the fifth ancestor of the current parental solution, following the rule:
\begin{equation}\label{eq:fifth-ancestor}
\bm{A} \leftarrow \sqrt{\alpha} \bm{A} + \frac{\sqrt{\alpha}}{\|\bm{z}\|^2}\left( \sqrt{1 - \frac{\beta}{\alpha}\|\bm{z}\|^2} - 1 \right) \bm{Az}\bm{z}^T
\end{equation}
where in this case $\alpha = \sqrt{1+c_{cov}^-}$ and $\beta = c_{cov}^-$, with $c_{cov}^-$ defined as in Table \ref{tab:oneplusoneparameters}.

In the specific case of constrained optimization, Arnold and Hansen \cite{Arnold2012} applied a similar rule as that of Equation \ref{eq:fifth-ancestor} also for decreasing the variance in the direction of constraint violations. The vector of directions of constraint violations $\bm{v}_j \leftarrow (1-c_c)\bm{v}_j + c_c\bm{Az}$ is updated at every constraint $j$ violated by step $\bm{Az}$, with $j \in 1,2,\ldots, m$. The rule is applied when at least one constraint is violated, updating the covariance matrix according to:
\begin{equation}
\bm{A} \leftarrow \bm{A} - \frac{B}{\sum_{j=1}^{m}\mathds{1}_{g_j(\bm{y}) > 0}} \sum_{j=1}^{m} \mathds{1}_{g_j(\bm{y}) > 0} \frac{\bm{v}_j\bm{w}_j^T}{\bm{w}_j\bm{w}_i^T}
\end{equation}
where $\bm{w_j} = \bm{A}^{-1}\bm{v_j}$.

\begin{table}[!ht]
\normalsize
\caption{Parameters of the (1+1)-CMA-ES with active covariance matrix updates \cite{Arnold2012}} \label{tab:oneplusoneparameters}
\begin{center}
\begin{tabular}{|ccc|} 
\hline
& & \\
 $c = \frac{2}{n+2}$ & $c_c = \frac{1}{n + 2}$ & $c_p = \frac{1}{12}$ \\
 & & \\
 $d = 1 + \frac{n}{2}$ & $B = \frac{0.1}{n + 2}$ & $c_{cov}^+ = \frac{2}{n^2 + 6}$ \\
 & & \\
$P_{thresh} = 0.44$ & $P_{target} = \frac{2}{11}$ & $c_{cov}^- = \frac{0.4}{n^{1.6} + 1}$ \\
& & \\
\hline
\end{tabular}
\end{center}
\end{table}
\vspace{-0.1cm}

\subsection{Viability Evolution}
Viability Evolution \cite{Mattiussi2003,maesani2014artificial} abstracts artificial evolutionary processes as a set of individuals, or candidate solutions, that must satisfy a number of viability criteria for surviving ever-changing environmental conditions. Viability criteria are defined as ranges of admissible values on problem objectives and constraints, the so-called viability boundaries. These boundaries, representing environmental conditions, are adapted during the evolutionary process to drive the evolving population towards desired regions of search space. At the beginning of the search the boundaries are relaxed to encompass all randomly generated initial solutions. Then, the boundaries are gradually tightened. Once viability boundaries reach the desired target, boundaries are not tightened further, and the evolutionary process is considered complete. 

Although in its simplest implementation the Viability Evolution paradigm operates by eliminating non-viable individuals, the abstraction is fully compatible with classical competition-based evolutionary algorithms. Elimination by viability and competition by objective functions can be modelled at the same time in the viability framework to determine the fitness of individuals. However, it is important to note that here fitness is intended as \emph{a-posteriori} reproduction capability and not as an \emph{a-priori} measurable function as currently implemented in the evolutionary computation practice.

Interestingly, some of the methods for constrained optimization, such as $\varepsilon$-DE\cite{Takahama:COb:cec2006}, ASCHEA\cite{Hamida02} and the constraint adaptation approach \cite{storn1999system}, can be loosely seen under the Viability Evolution abstraction. More specifically, $\varepsilon$-DE defines a tolerance $\varepsilon$ for comparing individuals by objective or constraints. Somehow, such a threshold could be seen as a viability boundary defined on the constraint violation. However, to be fully compatible with the viability paradigm, $\varepsilon$-DE should discard solutions violating the $\varepsilon$ tolerance on the constraint violations. Similarly, ASCHEA dynamically changes a tolerance on equality constraints for driving solutions towards the feasibility region determined by the equality constraint. In a similar way, the constraint adaptation approach progressively shrinks the feasible region during the evolutionary process.

\subsection{Introducing viability principles in CMA-ES and new rules for the adaptation of step-size: (1+1)-ViE-CMA-ES}\label{sec:viability-in-cma}

The definition of viability boundaries, which determine admissible (viable) regions of the search space at each generation, combined with the active covariance matrix updates for constrained optimization proposed in \cite{Arnold2012}, allows the search to be driven towards feasible areas of the search space by updating those boundaries. In \cite{viePPSN14}, we proposed a modified version of (1+1)-CMA-ES that is based on this idea. The resulting viability evolutionary algorithm, named (1+1)-ViE, makes use of a simple rule to update the viability boundaries, relax them to encompass the initial solutions at the first generation and tighten them during evolution, as better solutions, closer to the feasible region, are generated. 

Each constraint $g_j(\bm{x}) \leq 0, j = 1,2,\ldots,m$ is associated to a boundary $\bm{b}_j$, that is updated according to: 
\begin{align}
\bm{b}_{j} \leftarrow max\Big(0, min\Big(b_j, g_j(\bm{y}) + \frac{b_j-g_j(\bm{y})}{2}\Big)\Big)
\end{align}
whenever a successful offspring solution $\bm{y}$ is generated.

Furthermore, we extended the method presented in \cite{Arnold2012} to collect independent probabilities $p_{succ,j}$ of
success for each constraint $j$. These probabilities allow the determination of which boundary is more likely to be violated. We then reduced or increased the global $P_{succ}$ probability, used to adapt the step-size $\sigma$, following a simple heuristic: if the probability of success is lower than 50\% for at least one boundary, the step-size should be reduced (as per Equation \ref{ref:step-size-adaptation}). The $P_{succ}$ probability is in this case reduced as:
\begin{equation}
P_{succ} \leftarrow (1-c_p)P_{succ}.
\end{equation}
Instead, whenever a viable solution is generated, the $P_{succ}$ and $p_{succ,j}$ probabilities are increased as follows:
\begin{eqnarray}
P_{succ} \leftarrow (1-c_p)P_{succ} + c_p \\
p_{succ,j} \leftarrow (1-c_p)p_{succ,j} + c_p
\end{eqnarray}

In \cite{Arnold2012}, not adapting $P_{succ}$ on failure can lead to the use of outdated information for the step-size adaptation, whereas our heuristic tries to maintain and use only the most updated information, resulting in consistent performance gains on a set of unimodal constrained test functions \cite{viePPSN14}.
\section{Memetic Viability Evolution (\algorithmname)}\label{sec:algorithm}

Although the (1+1)-ViE-CMA-ES \cite{viePPSN14} showed good performance on unimodal constrained optimization problems, its capabilities of exploring the search space are limited by the use of a single offspring solution that is sampled at each iteration. This may strongly restrict its applicability on multi-modal landscapes or when disconnected feasible areas are present. 

\begin{figure}[!ht]
\centering
\includegraphics{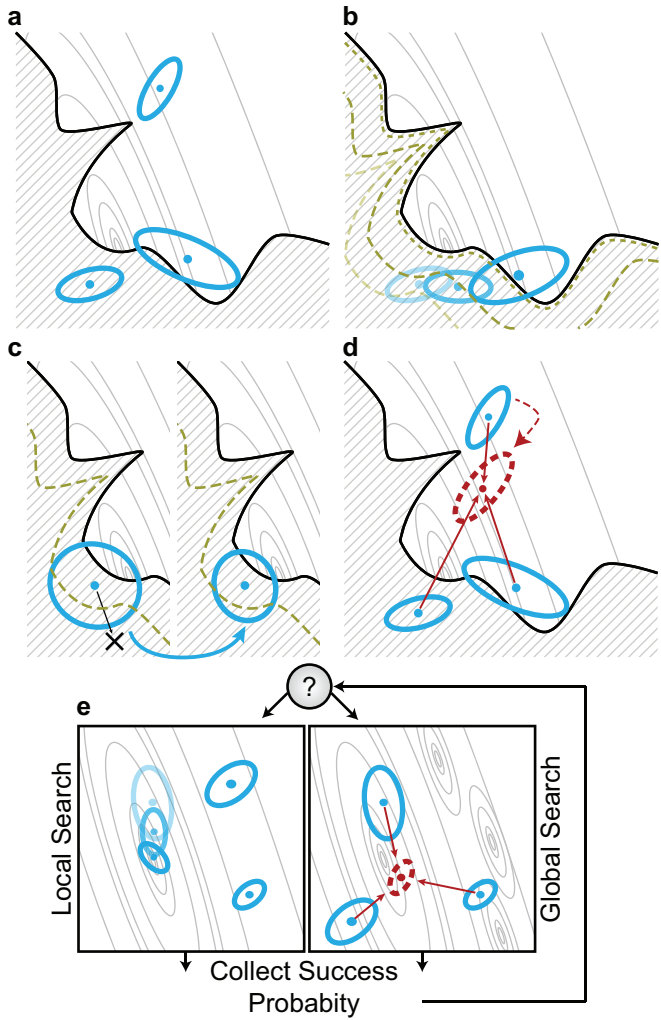}
\caption{Graphical representation of the main features of \algorithmname~on a simplified two-dimensional search landscape. Non-linear constraints on the search domain are represented as solid black lines, which outline the boundaries of the infeasible area marked with oblique gray hatching. Objective function contour lines are represented in the feasible areas as thin gray lines. Panels \textbf{a}, \textbf{b} and \textbf{c} represents the local search component of \algorithmname, while panel \textbf{d} the global search one. {\bf a}) Multiple search units (solid cyan ellipsoids, gray in print) sample solutions from different areas of the search landscape. A single search distribution is represented as an ellipsoid with its mean shown as a dot. {\bf b}) Constraints can be described as viability boundaries (dashed gold lines, gray in print) and can be relaxed. By progressively tightening the boundaries, while adapting the search distribution to the viability boundary violations, it is possible to drive the search units within the feasible areas.
{\bf c}) When a search distribution samples a solution that violates the viability boundaries (left), the most-likely direction of boundary violation is learned and used to adapt the search distribution to reduce the likelihood of searching in that direction (right). 
{\bf d}) Global search component. A new local search unit (dashed red ellipsoid, dark gray in print) is created by applying Differential Evolution mutation and crossover operators (solid arrows) on multiple search distributions. The parameters of the new distribution are inherited from the closest of the recombined search units (dashed arrow).
{\bf e}) The probability of improving solutions generated by local or global search operators are collected and used to adapt the activation frequencies of global or local search, either stepping a local search unit (left) or recombining information using a global search operator (right).
}
\label{fig:AlgoComponents}
\end{figure}

To overcome these limitations, we introduce here a memetic computing approach, called \algorithmname, that couples multiple local search units, based on (1+1)-ViE, with a global search operator used to recombine locally learned information. The multiple local search units can independently explore the constrained landscape, as illustrated in Fig. \ref{fig:AlgoComponents}a. When local search units sample solutions that violate viability boundaries, defined on the problem constraints, the search unit learns the direction of violation of the boundary \cite{Arnold2012} and adapts covariance matrix and step-size, as shown in Fig. \ref{fig:AlgoComponents}b. The adaptation of the viability boundaries, tightened during the search until they match the actual problem constraints, coupled with covariance matrix and step-size adaptation (as seen in Section \ref{sec:viability-in-cma}), drives the local search units towards feasible areas (Fig. \ref{fig:AlgoComponents}c). To enable global search, we combine local search units by using operators inherited from Differential Evolution. We recombine the mean of the search distributions of the (1+1)-ViE-CMA-ES units by using rand/1 mutation and exponential crossover \cite{book:price2005}. The search parameters of the new units (i.e., all the CMA-ES internal variables) are inherited from the closest of the search units participating in the rand/1 mutation, as illustrated in Fig. \ref{fig:AlgoComponents}d. However, if the closest search unit has converged, the parameters of the offspring unit are reinitialized to their default values. Crucially for the performance of the method, an adaptive scheduler adjusts the allocation of function evaluations to local search units or to global search operators (Fig. \ref{fig:AlgoComponents}e).

\begin{figure*}[!ht]%[H]
\centering
\includegraphics[width=\textwidth]{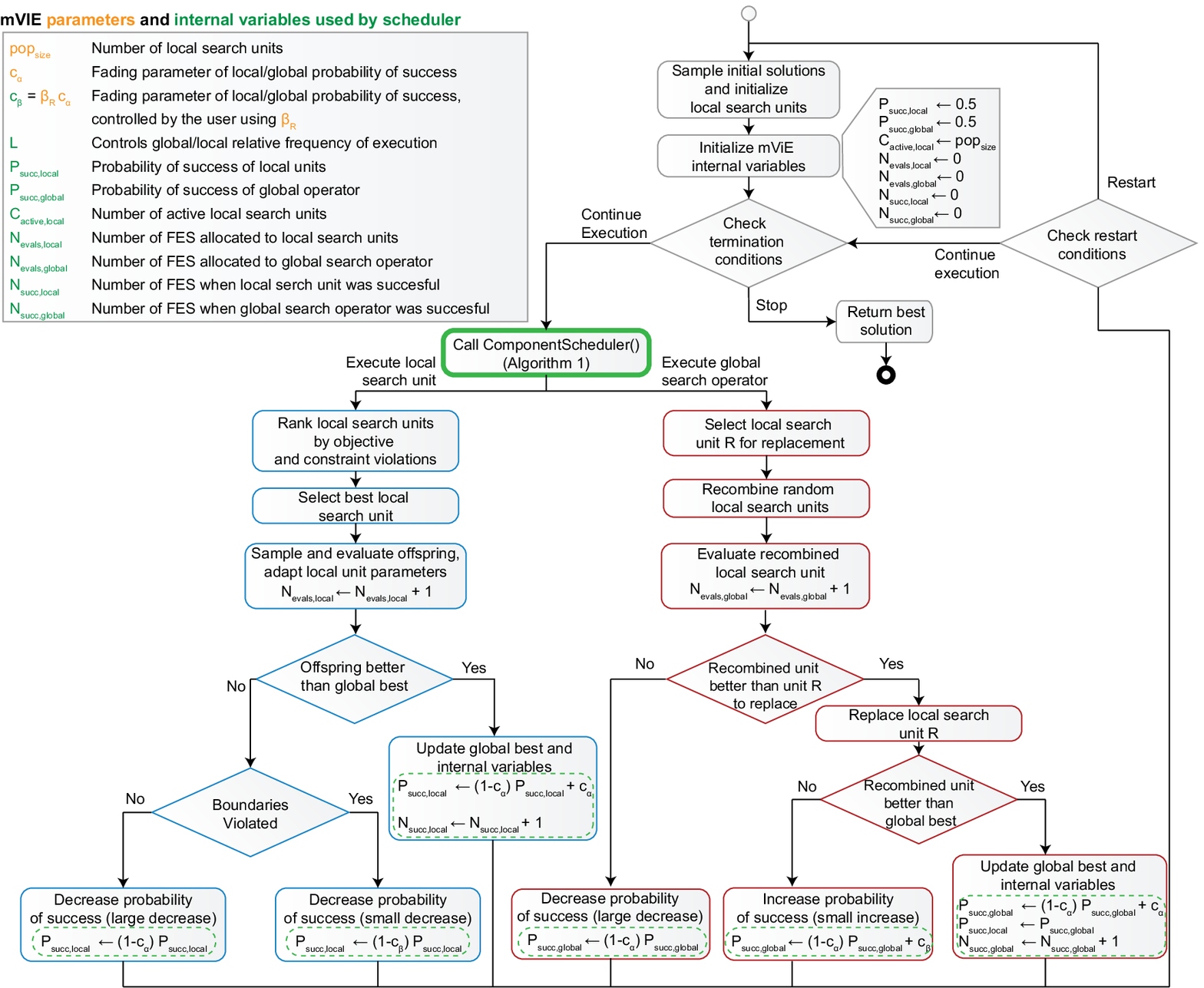}
\caption{Flowchart of \algorithmname. The algorithm's main operation can be divided in three main functional blocks. The scheduler (block with thick green border, gray in print) allocates a function evaluation either to a local search unit (left branch) or to the global search operator (right branch). Statistics used by the scheduler are collected after the execution of the two branches (dashed boxes).}
\label{fig:AlgoSchematics}
\end{figure*}

The flowchart of our method, implementing the operations described in Fig. \ref{fig:AlgoComponents}, is presented in Fig. \ref{fig:AlgoSchematics}. At the beginning of execution, \algorithmname~samples the desired number $pop_{size}$ of local search units and initializes internal variables that are used by the scheduler during the execution. At each iteration, the scheduler (Fig. \ref{fig:AlgoSchematics}, block with thick green border, gray in print) allocates a function evaluation either to a local search unit (Fig. \ref{fig:AlgoSchematics}, left branch) or to the global recombination operator (Fig. \ref{fig:AlgoSchematics}, right branch).

If a local search step is executed, the most promising unit in the population is selected, by ranking the local search units by fitness and constraints violation and selecting the first active (non-converged) top-ranking search unit. The (1+1)-ViE local search unit is then allowed to perform a single function evaluation. Finally, the moving average $P_{succ,local}$, i.e. the success probability of the local search units at improving the global best solution, is updated. To compare new candidate solutions with the current global best solution found, we employ the three feasibility rules presented in \cite{Deb2000}. Each local search unit maintains its viability boundaries and adapts them every time a better solution is sampled. The boundary on the objective is updated only when operating in the feasible area. Additional information on the probability of generating solutions that satisfy each constraint is maintained by each local search unit and used for updating the step-size, as described in Section \ref{sec:viability-in-cma}. After every execution of a local search unit, the unit is checked for convergence and disabled if one of the convergence criteria is met (see Section \ref{sec:term-cond}). 

On the other hand, when the global search step is executed, the search unit to be replaced is selected by picking two search units at random and choosing the worse one. The selected local search unit is then replaced if a better one is generated by means of Differential Evolution operators (applied to three randomly chosen search units). %rand/1 mutation and exponential crossover.
% \newline

After the local or global search steps are executed, the method collects statistics on the performance of the executed branch (Fig. \ref{fig:AlgoSchematics}, variable updates marked by dashed boxes). More specifically, it collects the number of function evaluations $N_{evals,\{local,global\}}$ and the number of improvements to the global best solution $N_{succ,\{local,global\}}$ of each search step. To put into effect the different meanings of ``success" for the local and global search steps, i.e. to weigh the relative ''importance`` of an improvement obtained by the two steps, the moving averaged probabilities of success of local and global search components are regulated by two additional parameters, $c_\alpha$ and $c_\beta$. In particular, when a solution better than the global best is found by a local search unit, the local step's probability of success is increased according to:
\begin{equation} \label{eq:increaseProb}
P_{succ,local} \leftarrow (1-c_\alpha)\cdot P_{succ,local} + c_\alpha.
\end{equation}
On the contrary, when a local search unit samples a solution that does not improve the global best, the probability is decreased according to: 
\begin{equation} \label{eq:decreaseProb}
P_{succ,local} \leftarrow (1-c_\alpha)\cdot P_{succ,local}. 
\end{equation}
However, if the solution violates boundaries defined on some constraints, to account for the function evaluations needed for adapting to the current boundaries, we discount the original $c_\alpha$ coefficient by a factor $\beta_R$, replacing it with a reduced coefficient $c_\beta = \beta_R \cdot c_\alpha$.

As for the global search step, the probability $P_{succ,global}$ is increased in two conditions. When a solution better than the global best solution is found, the probability is modified with a rule similar to Equation \ref{eq:increaseProb}: 
\begin{equation} \label{eq:increaseProbGlob}
P_{succ,global} \leftarrow (1-c_\alpha)\cdot P_{succ,global} + c_\alpha.
\end{equation}
The probability is also increased when a local search unit better than the parent unit is discovered, although this second increase is lower (we use the same coefficient $c_\beta$). In the other cases, the probability is decreased similar to Equation \ref{eq:decreaseProb}:
\begin{equation} \label{eq:decreaseProbGlob}
P_{succ,global} \leftarrow (1-c_\alpha)\cdot P_{succ,global}. 
\end{equation}

\subsection{Scheduler for selection of local/global search operators}\label{sec:scheduler}

The efficient allocation of function evaluations to local (Fig. \ref{fig:AlgoComponents}a-c; Fig. \ref{fig:AlgoSchematics}, left branch) or global (Fig. \ref{fig:AlgoComponents}d; Fig. \ref{fig:AlgoSchematics}, right branch) search is ensured by the scheduler presented in Algorithm \ref{alg:ComponentSelection} (Fig. \ref{fig:AlgoComponents}e; Fig. \ref{fig:AlgoSchematics}, block with thick green border, gray in print). The scheduler uses information collected during the search process, i.e. the moving average probability of success $P_{succ,\{local,global\}}$, the total number of global best improvements $N_{succ,\{local,global\}}$ and the total number of function evaluations $N_{evals,\{local,global\}}$ allocated to the local and global search components. 

\begin{algorithm}[!ht]%[H]
\caption{Selection of local/global search operator}  
\label{alg:ComponentSelection} 
\begin{algorithmic}
\State
\Function{componentScheduler}{$N_{evals,local}, \newline \mbox{\ \ \ \ \ \ \ } N_{evals,global}, N_{succ,local}, N_{succ,global}, L, C_{active,local}$}
\State
\If {$(N_{evals,local} + N_{evals, global}) < 100 \cdot n$}
	\State \textbf{return} \Comment{Set for execution local and global search}
\EndIf
     	
\If{$N_{evals,local} = 0$}
	\State $P_{local} \leftarrow 0$
\Else
	\State $P_{local} \leftarrow P_{curr,local} \cdot \frac{N_{succ,local}}{N_{evals,local}}$ 
\EndIf  		     
\If{$N_{evals,global} = 0$}
	\State $P_{global} \leftarrow 0$ 
\Else
	\State $P_{global} \leftarrow P_{curr,global} \cdot \frac{N_{succ,global}}{N_{evals,global}}$ 
\EndIf

\State $P_1 \leftarrow max(P_{local}, L \cdot P_{global})$
\State $P_2 \leftarrow max(P_{global}, L \cdot P_{local})$

\State
\If{$\Call{rand}{\null} < \frac{P_1}{P_1 + P_2} \land C_{active,local} > 0$}
	\State \textbf{return} \Comment{Set for execution local search}	
\Else        
	\State \textbf{return} \Comment{Set for execution global search}	
\EndIf
\State

\EndFunction

\end{algorithmic}
\end{algorithm}

During the first $100 \times n$ function evaluations (empirically set during preliminary experiments) both components are always used to ensure an initial learning phase and avoid unbalancing the search towards local or global search because of initial evaluations. Then, the total probability of success $P_{\{local,global\}} = \frac{N_{succ,\{local,global\}}}{N_{evals,\{local,global\}}} $, which considers all the previous history of the evolutionary run, is computed. We aggregate this information about total probability of success with the information on the current probability of success ($P_{succ,\{local,global\}}$) by multiplication. 
To prevent the frequency of execution of one of the two components from falling to zero, therefore disabling the component until the end of the search process, we limit the minimum frequency of execution of one component to a relative fraction $L$ of the other component's frequency. Thus, the frequency $f$ of execution of the two components is always limited $\frac{L}{1+L} \geq f \geq \frac{1}{1+L}, 0 \geq L \geq 1$. In the case where all the local search units have converged, i.e. the number of active local search units $C_{active,local}$ is zero, the local search step is disabled and only global search is performed.

\subsection{Termination conditions}\label{sec:term-cond}

Our method uses both local convergence conditions to disable local search units, and global convergence conditions to restart the algorithm. We disable a local search unit according to the standard stopping criteria for (1+1)-ViE, namely when:
\begin{enumerate}
\item the evolution path $\bm{s}$ multiplied by the step-size $\sigma$ is smaller than $10^{-12}$;
\item the maximum diagonal element on the covariance matrix multiplied by the step-size $\sigma$ is larger than $10^8$;
\item the condition number of the covariance matrix is larger than $10^{14}$.
\end{enumerate}

After a local search unit has been disabled, it can only be replaced with another active local search unit by the global search operator. The algorithm is restarted when all the local search units have converged to the same solution. This check is performed by measuring the difference between the mean of the objective and constraint violations of the local search units and the best constraint violation objective of the best solution.
\section{Experimental Setup}\label{sec:setup}

\begin{figure}[!ht]
\centering
\includegraphics{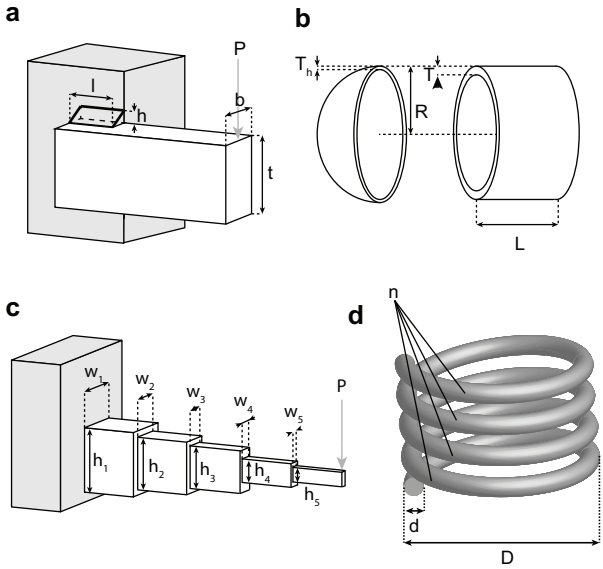}
\caption{Engineering problem benchmarks \cite{Coello2000,Gandomi11}. {\bf a}) Welded beam design optimization to minimize fabrication cost. The beam is fabricated out of carbon steel and welded on a rigid supporting structure. The shear force $P$ is loading the free tip of the beam. The dimensions of the beam (width $t$ and thickness $b$) and the width $h$ and length $l$ of the welded joint have to be optimized subject to constraints on shear stress, bending stress, buckling load on the bar, deflection and geometric constraints.
{\bf b}) Minimization of fabrication cost of a pressure vessel. The thickness of the spherical head $T_h$, the thickness of the spherical skin $T_s$, and the inner radius of the vessel have to be designed to comply with constraints derived from the ASME (American Society of Mechanical Engineers) standards on pressure vessels.
{\bf c}) Minimization of volume of a stepped cantilever. The cantilever is composed of five segments having variable cross-section, defined by the design variables $w_i$ and $h_i$. The design is subject to constraints limiting the bending stress and aspect ratio of each beam segment, and the total deflection of the cantilever at the tip.
{\bf d}) Design of a tension compression spring. The weight of the spring must be minimized optimizing mean coil diameter $D$, wire diameter $d$ and number of active coils $n$, subject to constraints on deflection, shear stress, surge frequency, and maximum size on the outside diameter.}
\label{fig:EngineeringProblems}
\end{figure}

To assess the performance of the proposed method, we selected all the optimization problems with only inequality constraints from the CEC 2006 benchmark\cite{ProbCEC2006}\footnote{Although a newer CEC benchmark is available \cite{ProbCEC2010}, with some interesting features such as scalable functions, the CEC 2006 benchmark still represents the most popular tool for testing new constrained optimization algorithms, given the availability of results for several methods from the literature.}. A summary of the tested benchmark functions, with their relative features (problem dimension, kinds of constraints, number of constraints active at the optimum) is reported in Table \ref{tab:CecProblems}\footnote{We should note that our algorithm is specifically designed for handling optimization problems with inequality constraints only. A more detailed analysis of the behaviour of \algorithmname~on problems including (also) equality constraints, together with the original MATLAB\textsuperscript{\textregistered} source code of all our experiments, is available at \url{http://lis.epfl.ch/mVIE}.} %\url{http://lis.epfl.ch/files/content/users/195419/files/mvie/index.html}
As specified in \cite{ProbCEC2006}, we executed our algorithm for 500000 function evaluations on each test function. For each function, we measured the success rate (SR) over 25 runs of \algorithmname~the number of function evaluations (NFES) needed for solving the problem at the desired accuracy (set to $10^{-4}$). Furthermore, we tested the method on a set of four engineering problems reported in Fig. \ref{fig:EngineeringProblems}. The engineering problems were run for a maximum of 200000 function evaluations.

In the experiments, we set the algorithm's main parameters after performing a preliminary parameter analysis, reported in Appendix \ref{apx:parameter-study}. The identified parameter values, used in the rest of this paper, are $c_\alpha = 0.1, \beta_R = 0.05, L = 0.18, pop_{size} = 40$. For the DE operators, we used the standard values reported in the literature \cite{book:price2005}, namely $F = 0.5$ and $CR = 0.9$. The parameters of the local search component, based on (1+1)-CMA-ES, are as set in \cite{Arnold2012}, reported in Table \ref{tab:oneplusoneparameters}. All the experiments were performed on Intel\textsuperscript{\textregistered} machines with Core\texttrademark~i7-2600 CPU $@$ 3.40GHz and 8GB of RAM.

\begin{table}[!ht]
\renewcommand{\arraystretch}{1.2}
%\caption{Characteristic of the 24 benchmark problems from the CEC 2006 competition on constrained optimization. We show the number of dimensions $n$, the number of linear equalities (LE) and inequalities (LI) and the number of non-linear equalities (NE) and inequalities (NI) for each test problem. Moreover, the number of active constraints at the optimum is reported (A). We used the check-marked problems in the experiments presented in this paper, which do not contain equality constraints.}
\caption{Problems with inequality constraints from the CEC 2006 competition on constrained optimization. We show the number of dimensions ($n$), linear (LI), non-linear (NI) and active constraints at the optimum ($a$).}
\label{tab:CecProblems}
\centering
%\begin{tabular}{|l|c|c|c|c|c|c|c|}
%\cline{2-8}
%\multicolumn{1}{c|}{} & Problem & n & LI & NI & LE & NE & a \\
%\cline{1-8}
%\checkmark & g01 & 13 &  9 & 0 & 0 & 0 & 6\\
%\checkmark & g02 & 20 &  0 & 2 & 0 & 0 & 1 \\
%& g03 & 10 & 0 & 0 & 0 & 1 & 1\\ 
%\checkmark & g04 & 5  & 0 & 6 & 0 & 0 & 2 \\
%& g05 & 4 &  2 & 0 & 0 & 3 & 3 \\
%\checkmark & g06 & 2  &  0 & 2 & 0 & 0 & 2 \\
%\checkmark & g07 & 10 & 3 & 5 & 0 & 0 & 6 \\
%\checkmark & g08 & 2  & 0 & 2 & 0 & 0 & 0 \\ 
%\checkmark & g09 & 7   & 0 & 4 & 0 & 0 & 2 \\
%\checkmark & g10 & 8   & 3 & 3 & 0 & 0 & 6 \\
%& g11 & 2   & 0 & 0 & 0 & 1 & 1 \\
%\checkmark & g12 & 3   & 0 & 1 & 0 & 0 & 0 \\ 
%& g13 & 5   & 0 & 0 & 0 & 3 & 3 \\ 
%& g14 & 10   & 0 & 0 & 3 & 0 & 3 \\
%& g15 & 3  & 0 & 0 & 1 & 1 & 2 \\
%\checkmark & g16 & 5   & 4 & 34 & 0 & 0 & 4 \\
%& g17 & 6   & 0 & 0 & 0 & 4 & 4 \\
%\checkmark & g18 & 9   & 0 & 13 & 0 & 0 & 6 \\
%\checkmark & g19 & 15  & 0 & 5 & 0 & 0 & 0 \\
%& g20 & 24  & 0 & 6 & 2 & 12 & 16 \\
%& g21 & 7  & 0 & 1 & 0 & 5 & 6 \\
%& g22 & 22  & 0 & 1 & 8 & 11 & 19 \\
%& g23 & 9 & 0 & 2 & 3 & 1 & 6 \\
%\checkmark & g24 & 2  & 0 & 2 & 0 & 0 & 2 \\ 
%\cline{1-8}
%\end{tabular}
\begin{tabular}{|c|c|c|c|c|c|c|}
\hline
Problem & n & LI & NI  & a \\
\hline
g01 & 13 &  9 & 0 &  6\\
g02 & 20 &  0 & 2 &  1 \\
g04 & 5  & 0 & 6 & 2 \\
g06 & 2  &  0 & 2 &  2 \\
g07 & 10 & 3 & 5 &  6 \\
g08 & 2  & 0 & 2 &  0 \\ 
g09 & 7   & 0 & 4 &  2 \\
g10 & 8   & 3 & 3 &  6 \\
g12 & 3   & 0 & 1 &  0 \\ 
g16 & 5   & 4 & 34 &  4 \\
g18 & 9   & 0 & 13 &  6 \\
g19 & 15  & 0 & 5 &  0 \\
g24 & 2  & 0 & 2 &  2 \\ 
\hline
\end{tabular}
\vspace{-1.0em}
\end{table}

\section{Results}\label{sec:results}
We first review the results on CEC 2006 problems, then we move forward to analyze the results on the engineering problems. Moreover, we show a more detailed analysis of few selected runs of \algorithmname, to exemplify its behaviour on different fitness landscapes. Finally, we dissect the performance of \algorithmname~by assessing the contribution of each of its components.

\subsection{CEC 2006 problems}

On all the 25 runs of every CEC 2006 test problem with only inequality constraints, our method exhibited 100\% success rate (SR), i.e. it could reach in every run the target fitness difference from the optimum ($10^{-4}$). The number of function evaluations (NFES) needed for reaching success is reported in Table \ref{tab:CecInequalitySummary}. Furthermore, in Appendix \ref{apx:error-values} we report the statistics on the error values, according to the CEC 2006 format \cite{ProbCEC2006}.

\begin{table}[!ht]
%\vspace{-1.4em}
\caption{Best, Median, Worst, Mean and Std. Dev. of NFES to achieve the fixed accuracy level ($(f(\bm{x})-f(\bm{x}^*)) \leq 0.0001$), 
and Success Rate over 25 runs of \algorithmname~on the selected CEC 2006 problems.} \label{tab:CecInequalitySummary}
{
%\scriptsize
\fontsize{7pt}{10pt}\selectfont
\setlength{\tabcolsep}{6pt}
\begin{center}
\begin{tabular}{|c|c|c|c|c|c|c|} \hline
{\bf Prob.}& {\bf Best}  & {\bf Median} & {\bf Worst} & {\bf Mean} & {\bf Std} & {\bf SR} \\
\hline g01 & 15032 & 20304 & 26301 & 20645.8 & 2757.39 & 100\%\\ 
\hline g02 & 42462 & 61072 & 200394 & 67972.4 & 30770.3 & 100\%\\ 
\hline g04 & 3089 & 3945 & 12461 & 4540.88 & 2154.32 & 100\%\\ 
\hline g06 & 1072 & 1901 & 7222 & 2782.84 & 1884.95 & 100\%\\ 
\hline g07 & 5954 & 7281 & 44035 & 10511.9 & 8732.84 & 100\%\\ 
\hline g08 & 185 & 482 & 812 & 504.2 & 183.513 & 100\%\\ 
\hline g09 & 2586 & 3436 & 26837 & 5193.68 & 5064.41 & 100\%\\ 
\hline g10 & 10995 & 14734 & 99587 & 23884.3 & 22860.9 & 100\%\\ 
\hline g12 & 187 & 3809 & 12607 & 3967.32 & 2371.44 & 100\%\\ 
\hline g16 & 2415 & 3128 & 10214 & 3843.12 & 2100.47 & 100\%\\ 
\hline g18 & 4683 & 7272 & 73350 & 13916.9 & 17785.5 & 100\%\\ 
\hline g19 & 22658 & 25914 & 35753 & 26770.1 & 3029.09 & 100\%\\ 
\hline g24 & 492 & 718 & 2511 & 838.68 & 400.114 & 100\%\\ 
\hline
\end{tabular}
\end{center}
}
\end{table}

We compared \algorithmname's SR and median NFES needed to solve the CEC 2006 problems against representative algorithms from the state-of-the-art in constrained optimization 
that \emph{do not} make use of traditional non-linear programming (NLP) techniques, see Table \ref{tab:ComparedAlgo} (top). For convenience, we grouped the algorithms under comparison based on their underlying meta-heuristic: DE, CMA-ES, PSO, and algorithms which are based on other evolutionary paradigms (GA, ES, and hybrid variants of ES/DE). The latter group is labeled as ``others''. DE-based and CMA-ES-based algorithms have been considered for comparison as DE and CMA-ES represent the building blocks of the proposed \algorithmname. Therefore, it is interesting to compare our method with other algorithms using similar logics. The two other groups (PSO-based methods and ``others'') were included to show how our method compares with other successful examples from the state-of-the-art that are inspired by different computational paradigms. Regardless of the underlying meta-heuristic, it is worth noting how each of the compared algorithms ranks feasible/infeasible solutions: with respect to this aspect, we can see from Table \ref{tab:ComparedAlgo} that the majority of the algorithms in the state-of-the-art use $\varepsilon$-ranking, three feasibility rules or (adaptive) penalty functions. Few methods use instead algorithm-specific methods, e.g. based on surrogate models or learning of the feasibility structure.

Table \ref{tab:CecComparison} shows the comparison with all the aforementioned algorithms. As an additional comparison, the table also includes the aggregate best results obtained on each problem by \emph{all} the algorithms presented at CEC 2006. For the reader's convenience, we report again (separately) the NFES and SR of \algorithmname. For each problem, we also report the relative difference of NFES (labeled as $\Delta$NFES\%) between \algorithmname~and the best algorithm (underlined) from each group of algorithms, i.e.:
\begin{center}
$\Delta \text{NFES}\%=\frac{\text{NFES}_{mViE}-\text{NFES}_{bestGroupAlg}}{\text{NFES}_{bestGroupAlg}}\times100$.
\end{center}
We calculate in a similar way the $\Delta$NFES\% w.r.t. the best results from CEC 2006. In the table, the symbol ``-'' indicates that the result of that algorithm for that problem is not available.

\begin{table*}
\caption{\algorithmname~ was compared against multiple algorithms selected from the literature: algorithms not using NLP techniques (top); memetic algorithms using NLP techniques (bottom). For each algorithm we summarize the main features, together with the method used by the algorithm to compare solutions and handle constraints.} \label{tab:ComparedAlgo}
\begin{center}
\begin{tabular}{|p{0.3cm}|m{2.3cm}|>{\raggedright}m{3.2cm}|m{11.2cm}|}
\hline & \textbf{Algorithm} & \textbf{Ranking method} & \textbf{Notes}\\
\hline & Eps-DE \cite{Takahama:COb:cec2006} & $\varepsilon$-ranking & Gradient-based mutations.\\
\cline{2-4} & Eps-RDE \cite{DBLP:conf/cec/TakahamaS12a} & $\varepsilon$-ranking & Surrogate function is used in $\varepsilon$-ranking comparisons.\\
\cline{2-4} & MPDE \cite{Tasgetiren:AMD:cec2006} & Penalty Function (Near Feasibility Threshold \cite{Smith1993}) & Uses multiple sub-populations to maintain diversity. \\
\cline{2-4} & GDE \cite{Kukkonen06} & Non dominated sorting & Problem reformulated as multi-objective problem using sum of constraint violations as objective.\\
\cline{2-4} & MDE \cite{Mezura-Montes:MDE:cec2006} & Three feasibility rules\textsuperscript{A} &  Modified DE mutation operator that considers the best and three other randomly selected individuals. Uses a diversity procedure based on stochastic ranking with probability modified during search.\\
\cline{2-4} \multirow{-8}{*}{\begin{sideways}DE\end{sideways}} & JDE-2 \cite{Brest:SDE:cec2006} & Three feasibility rules\textsuperscript{B} & Adapts the F and CR parameters of DE during search.\\
\hline %\Xhline{4\arrayrulewidth} \parbox[t]{2mm}{\multirow{6}{*}{\rotatebox[origin=c]{90}{DE}}} 
       & (1+1)-aCMA \cite{Arnold2012} & If offspring violates constraints, adapts covariance matrix, otherwise substitutes parent if better fitness. &\\
\cline{2-4} & (1+1)-ViE \cite{viePPSN14} & As in (1+1)-aCMA & Includes a mechanism for adapting the step-size based on information gathered on constraint violations. Relaxes constraints and uses them to drive the search towards feasible area.\\
\cline{2-4} & APM-CMA-ES \cite{Kusakci2013a} & Adaptive penalty function & Adapts tolerances on equality constraints during the search.\\
\cline{2-4} \multirow{-8}{*}{\begin{sideways}{   CMA-ES}\end{sideways}} & AP-CMA-ES \cite{deMelo2014} & Adaptive penalty function & Aggregates the violation of all solutions in the current population.\\
\hline %\Xhline{4\arrayrulewidth} \parbox[t]{2mm}{\multirow{4}{*}{\rotatebox[origin=c]{90}{CMA-ES}}} 
       & PSO \cite{Zielinski:CSO-b:cec2006} & Three feasibility rules\textsuperscript{A} &\\
\cline{2-4} & COPSO \cite{Aguirre:2007} & Three feasibility rules\textsuperscript{A} & Adapts tolerances on equality constraints during the search. Maintains an archive of solutions that are estimated to be close to constraint boundary. Applies local perturbation of best solutions found.\\
\cline{2-4} \multirow{-4}{*}{\begin{sideways}PSO\end{sideways}} & PESO+ \cite{Munoz-Zavala:CSO:cec2006} & Three feasibility rules\textsuperscript{A} & Adapts tolerances on equality constraints during the search. Maintains an archive of solutions close to constraint boundary. Applies local perturbation of best solutions found.\\
\hline %\Xhline{4\arrayrulewidth} \parbox[t]{2mm}{\multirow{3}{*}{\rotatebox[origin=c]{90}{PSO}}} 
        & ASRES \cite{Runarsson:AES:cec2006} & Stochastic ranking performed using surrogate models of constraints and objectives. & ES\\
\cline{2-4} & ($\mu$+$\lambda$)-CDE \cite{Wang11a} & Adaptive trade-off model to select offspring for next generation that depends on population composition (only feasible, only infeasible or mixed). & ($\mu$+$\lambda$)-DE. Selects adaptively among various DE mutation strategies.\\
\cline{2-4} & ICDE \cite{Jia2013302} & As in ($\mu$+$\lambda$)-CDE & ($\mu$+$\lambda$)-DE. Uses an archiving strategy in the infeasible population case.\\
\cline{2-4} \multirow{-11}{*}{\begin{sideways}{    Others}\end{sideways}} & PCX \cite{Sinha:APP:cec2006} & Three feasibility rules\textsuperscript{C} & Steady-state GA with PCX recombination operator. Adapts tolerances on equality constraints during the search.\\
%\hline %\Xhline{4\arrayrulewidth} \parbox[t]{2mm}{\multirow{4}{*}{\rotatebox[origin=c]{90}{Others}}}
\hline \multicolumn{4}{|p{\textwidth}|}{
\vspace{1pt}
\textsuperscript{A} the error for infeasible solutions is computed as the sum of constraint violations \newline
\textsuperscript{B} the error for infeasible solutions is computed as the mean of the constraint violations \newline
\textsuperscript{C} the error for infeasible solutions is computed as the sum of constraint violations, the fitness is re-scaled according to special rules \newline
\vspace{1pt}
}\\
\hline
\end{tabular}

\vspace{0.5cm}

\begin{tabular}{|p{0.3cm}|m{2.3cm}|>{\raggedright}m{3.2cm}|m{11.2cm}|}
\hline & \textbf{Algorithm} & \textbf{Ranking method} & \textbf{Notes}\\
\hline & SADE \cite{Huang:SDE:cec2006} & Three feasibility rules\textsuperscript{A} & \mbox{DE +} SQP. Selects probabilistically among various DE mutation strategies.\\
\cline{2-4} & DMS-PSO \cite{Liang:DMP:cec2006}& Uses one constraint (adaptively chosen) for each sub-population as objective. & \mbox{PSO +} SQP. Maintains multiple sub-populations.\\
\cline{2-4} & GB-MA \cite{SunGaribaldi2010} & Over-penalize approach \cite{Runarsson1993}: ranks first feasible solutions by objective, then infeasible ones by sum of constraint violations. & \mbox{EDA +} DONLP2\\
\cline{2-4} \multirow{-11}{*}{\begin{sideways}{  MA with NLP Techniques}\end{sideways}} & FSM \cite{Handoko2010} & Neighborhood-modulated selection & \mbox{GA +} SQP. Maintains a database of evaluated solutions. Uses Support Vector Machine to model the structure of the feasible region and choose which solutions are significant for local refinements.\\
% \parbox[t]{2mm}{\multirow{4}{*}{\rotatebox[origin=c]{90}{EA+NLP Techniques}}} 
\hline \multicolumn{4}{|p{\textwidth}|}{
\vspace{1pt}
%\textsuperscript{A} the error for infeasible solutions is computed as the sum of constraint violations \newline
%\textsuperscript{B} the error for infeasible solutions is computed as the mean of the constraint violations \newline
%\textsuperscript{C} the error for infeasible solutions is computed as the sum of constraint violations, the fitness is rescaled according to special rules \newline
\textsuperscript{A} the error for infeasible solutions is  computed as the mean of constraint violations, normalized by the maximum violation observed for each constraint \newline
\vspace{1pt}
} \\
\hline
\end{tabular}
\end{center}
\end{table*}
\begin{table*}[!ht]
\caption{Median NFES to achieve the fixed accuracy level ($(f(\bm{x})-f(\bm{x}^*)) \leq 0.0001$) and Success Rate for the selected CEC 2006 problems. All the compared results are 
obtained from the corresponding papers. Please note that results reported for Eps-RDE \cite{DBLP:conf/cec/TakahamaS12a} and PCX\cite{Sinha:APP:cec2006} are, respectively, mean and $25^{th}$ percentile, rather than median NFES. BEST CEC 2006 aggregates the best results obtained on each problem by all the algorithms presented at CEC 2006. 
$\Delta$NFES\% indicates for each problem the relative difference of NFES between the proposed algorithm \algorithmname~and the best algorithm in each group (underlined).} \label{tab:CecComparison}
{
%\tiny
%\fontsize{5.5pt}{6pt}\selectfont
\fontsize{7pt}{10pt}\selectfont
\setlength{\tabcolsep}{2pt}
\begin{center}
\begin{tabular}{|c|c|c|} \hline
\multicolumn{3}{|c|}{\textbf{This work}} \\ 
\hline
\multirow{2}{*}{\textbf{Prob.}} & \multicolumn{2}{c|}{\textbf{\algorithmname  }} \\ 
%\multirow{3}{*}{\textbf{Prob.}} & \multicolumn{2}{c|}{\multirow{2}{*}{\textbf{\algorithmname  }}} \\ 
%& \multicolumn{2}{c|}{}\\ 
\cline{2-3}
	    & \textbf{NFES} & \textbf{SR}\\ 
\hline g01  &  {20304}  &  100\% \\ 
\hline g02  &  {61072}  &  {100\%} \\ 
\hline g04  &  {3945}  &  100\% \\ 
\hline g06  &  1901  &  100\% \\ 
\hline g07  &  {7281}  & 100\% \\ 
\hline g08  &  482  &  100\% \\ 
\hline g09  &  {3436}  & 100\% \\ 
\hline g10  &  14734  &  100\% \\ 
\hline g12  &  {3809}  &  100\% \\ 
\hline g16  &  {3128}  &  100\% \\ 
\hline g18  &  {7272}  &  100\% \\ 
\hline g19  &  {25914}  &  100\% \\ 
\hline g24  &  {718}  &  100\% \\ 
\hline
\end{tabular}
\hspace{0.3cm}
\begin{tabular}{|c|c|c|c|c|c|c|c|c|c|c|c|c|c|c|c|} \hline
\multicolumn{14}{|c|}{\textbf{DE}} \\
\hline
\multirow{2}{*}{\textbf{Prob.}} & \multicolumn{2}{c|}{\textbf{Eps-DE \cite{Takahama:COb:cec2006} }} & \multicolumn{2}{c|}{\textbf{Eps-RDE \cite{DBLP:conf/cec/TakahamaS12a} }} & \multicolumn{2}{c|}{\textbf{MPDE \cite{Tasgetiren:AMD:cec2006} }} & \multicolumn{2}{c|}{\textbf{GDE \cite{Kukkonen06} }} & \multicolumn{2}{c|}{\textbf{MDE \cite{Mezura-Montes:MDE:cec2006} }} & \multicolumn{2}{c|}{\textbf{jDE-2 \cite{Brest:SDE:cec2006} }} & \multirow{2}{*}{\textbf{$\Delta$NFES\%}} \\  
\cline{2-13}
& \textbf{NFES} & \textbf{SR}  & \textbf{NFES} & \textbf{SR}  & \textbf{NFES} & \textbf{SR}  & \textbf{NFES} & \textbf{SR}  & \textbf{NFES} & \textbf{SR}  & \textbf{NFES} & \textbf{SR} & \\
\hline  g01  &  59345  &  100\%  &  56508  &  100\%  &  43794  &  100\%  &  \underline{40200}  &  100\%  &  75000  &  100\%  &  50354  &  100\% &-49.49\%\\ 
\hline  g02  &  146911  &  100\%  &  99742  &  100\%  &  280272  &  100\%  &  106332  &  72\%  &  \underline{71100}  &  16\% & 138102  &  92\% &-14.10\%\\ 
\hline  g04  &  26098  &  100\%  &  51614  &  100\%  &  20823  &  100\%  &  \underline{15157}  &  100\%  &  39300  &  100\% &  40958  &  100\% &-73.97\%\\ 
\hline  g06  &  7316  &  100\%  &  10152  &  100\%  &  10550  &  100\%  &  6431  &  100\%  &  \underline{5250}  &  100\%  &  29844  &  100\% &-63.79\%\\ 
\hline  g07  &  74476  &  100\%  &  99830  &  100\%  &  \underline{57079}  &  100\%  &  112969  &  100\%  &  176400  &  100\%  &  126637  &  100\% &-87.24\%\\ 
\hline  g08  &  1182  &  100\%  &  4063  &  100\%  &  1632  &  100\%  &  1486  &  100\%  &  \underline{900}  &  100\%  &  3564  &  100\% &-46.44\%\\ 
\hline  g09  &  23172  &  100\%  &  42266  &  100\%  &  20814  &  100\%  &  30784  &  100\%  &  \underline{15000}  &  100\% &  55515  &  100\% &-77.09\%\\ 
\hline  g10  &  105799  &  100\%  &  99820  &  100\%  &  \underline{48508}  &  100\%  &  81827  &  100\%  &  163500  &  100\%  &  144247  &  100\% &-69.62\%\\ 
\hline  g12  &  4155  &  100\%  &  7873  &  100\%  &  4227  &  100\%  &  3016  &  100\%  &  \underline{1200}  &  100\%  &  6684  &  100\% & +217.41\%\\ 
\hline  g16  &  13001  &  100\%  &  - &  -  &  13135  &  100\%  &  13307  &  100\%  &  \underline{8700}  &  100\%  &  261549  &  100\% & -64.04\%\\ 
\hline  g18  &  59232  &  100\%  &  - &  -  &  \underline{42550}  &  100\%  &  377732  &  76\%  &  118050  &  100\%  &  449306  &  100\% &-82.90\%\\ 
\hline  g19  &  354060  &  100\%  &  - &  -  &  115054  &  100\%  &  206556  &  88\%  &  - &  -  & \underline{101076}  &  100\% &-74.36\%\\ 
\hline  g24  &  2928  &  100\%  &  - &  -  &  4371  &  100\%  &  3059  &  100\%  &  \underline{1650}  &  100\%  &  319611  &  100\% &-56.48\%\\
\hline
\end{tabular}

\vspace{0.25cm} 

\begin{tabular}{|c|c|c|c|c|c|c|c|c|c|} \hline
\multicolumn{10}{|c|}{\textbf{CMA-ES}} \\ \hline
\multirow{2}{*}{\textbf{Prob.}} & \multicolumn{2}{c|}{\textbf{(1+1)-aCMA \cite{Arnold2012} }} & \multicolumn{2}{c|}{\textbf{(1+1)-VIE \cite{viePPSN14} }} & \multicolumn{2}{c|}{\textbf{APM-CMA-ES \cite{Kusakci2013a} }} & \multicolumn{2}{c|}{\textbf{AP-CMA-ES \cite{deMelo2014} }} & \multirow{2}{*}{\textbf{$\Delta$NFES\%}} \\ 
\cline{2-9}
            & \textbf{NFES} & \textbf{SR}  & \textbf{NFES} & \textbf{SR}  & \textbf{NFES} & \textbf{SR}  & \textbf{NFES} & \textbf{SR} & \\ 
\hline g01  & - &  -  &  - &  -  &  \underline{51400}  &  100\%  &  184778  &  52\% & -60.50\%\\ 
\hline g02  & - &  -  &  - &  -  &  \underline{1328100}  &  30\%  &  - &  - & -95.40\%\\ 
\hline g04  & - &  -  &  - &  -  &  25700  &  100\%  &  \underline{4896}  &  100\% & -19.42\%\\ 
\hline g06  & 1060  &  100\%  &  \underline{900}  &  100\%  &  7300  &  100\%  &  2424  &  100\% & +111.22\%\\ 
\hline g07  & 11283  &  100\%  &  \underline{7545}  &  100\%  &  116800  &  100\%  &  14420  &  100\% & -3.50\%\\ 
\hline g08  & - &  -  &  - &  -  &  1500  &  100\%  &  \underline{348}  &  100\% & +38.50\%\\ 
\hline g09  & 4106  &  100\%  &  \underline{3660}  &  100\%  &  77400  &  100\%  &  5346  &  100\% & -6.12\%\\ 
\hline g10  & 18781  &  100\%  &  \underline{8295}  &  100\%  &  407400  &  100\%  &  23780  &  100\% & +77.62\%\\ 
\hline g12  & - &  -  &  - &  -  &  \underline{7500}  &  100\%  &  26278  &  100\% & -49.21\%\\ 
\hline g16  & - &  -  &  - &  -  &  - &  -  &  \underline{5648}  &  100\% & -44.62\%\\ 
\hline g18  & - &  -  &  - &  -  &  - &  -  &  \underline{57430}  &  100\% & -87.34\%\\ 
\hline g19  & - &  -  &  - &  -  &  - &  -  &  \underline{74472}  &  100\% & -65.20\%\\ 
\hline g24  & - &  -  &  - &  -  &  - &  -  &  \underline{996}  &  100\% & -27.91\%\\ 
\hline
\end{tabular}
\hspace{0.3cm}
\begin{tabular}{|c|c|c|c|c|c|c|c|} \hline
\multicolumn{8}{|c|}{\textbf{PSO}} \\ \hline
\multirow{2}{*}{\textbf{Prob.}} & \multicolumn{2}{c|}{\textbf{PSO \cite{Zielinski:CSO-b:cec2006} }} & \multicolumn{2}{c|}{\textbf{COPSO \cite{Aguirre:2007} }} & \multicolumn{2}{c|}{\textbf{PESO+ \cite{Munoz-Zavala:CSO:cec2006} }} & \multirow{2}{*}{\textbf{$\Delta$NFES\%}} \\ 
\cline{2-7}
& \textbf{NFES} & \textbf{SR}  & \textbf{NFES} & \textbf{SR}  & \textbf{NFES} & \textbf{SR} & \\ 
\hline g01  & \underline{46405}  &  52\%  &  95000  &  30\%  &    102100  &  100\% & -56.24\%\\ 
\hline g02  &  - &  -  &  \underline{175800}  &  22\%  &  219400  &  56\% & -65.26\%\\ 
\hline g04  &  \underline{19681}  &  100\%  &  65100  &  30\%  &  79300  &  100\% & -79.95\%\\ 
\hline g06  &  \underline{20007}  &  100\%  &  54200  &  30\%  &  56800  &  100\% & -90.50\%\\ 
\hline g07  &  327283  &  8\%  &  \underline{227600}  &  30\%  &  358600  &  96\% & -96.80\%\\ 
\hline g08  &  \underline{2311}  &  100\%  &  6850  &  30\%   &  6100  &  100\% & -79.14\%\\ 
\hline g09  &  \underline{57690}  &  100\%  &  78500  &  30\%  &  96400  &  100\% & -94.04\%\\ 
\hline g10  &  461422  &  32\%  &  \underline{221300}  &  30\%   &  468350  &  16\% & -93.34\%\\ 
\hline g12  &  \underline{3933}  &  100\%  &  6900  &  30\%  &  8100  &  100\% & -3.15\%\\ 
\hline g16  &  \underline{33021}  &  100\%  &  41000  &  30\%  &  48700  &  100\% & -90.52\%\\ 
\hline g18  &  177989  &  80\%  &  \underline{153600}  &  27\%  &  211800  &  92\% & -95.26\%\\ 
\hline g19  &  365284  &  8\%  &  \underline{259650}  &  14\%  &  - &  - & -90.02\%\\ 
\hline g24  &  \underline{7487}  &  100\%  &  19350  &  30\%  &  19900  &  100\% & -90.41\%\\ 
\hline
\end{tabular}

\vspace{0.25cm} 

\renewcommand{\arraystretch}{1.008}
\begin{tabular}{|c|c|c|c|c|c|c|c|c|c|} \hline
\multicolumn{10}{|c|}{\textbf{Others}} \\ 
\hline
\multirow{2}{*}{\textbf{Prob.}} & \multicolumn{2}{c|}{\textbf{ASRES \cite{Runarsson:AES:cec2006} }} & \multicolumn{2}{c|}{\textbf{($\mathbf{\mu}$+$\mathbf{\lambda}$)-CDE \cite{Wang11a} }} & \multicolumn{2}{c|}{\textbf{ICDE \cite{Jia2013302} }} & \multicolumn{2}{c|}{\textbf{PCX \cite{Sinha:APP:cec2006} }} & \multirow{2}{*}{\textbf{$\Delta$NFES\%}} \\ 
\cline{2-9}
 & \textbf{NFES} & \textbf{SR}  & \textbf{NFES} & \textbf{SR}  & \textbf{NFES} & \textbf{SR} & \textbf{NFES} & \textbf{SR} & \\ 
\hline g01  & 62800  &  100\%  & 89320  &  100\%  &  106540  &  100\%  &  \underline{62026}  &  100\%  & -67.26\%\\
\hline g02  & 321200  &  12\%  &  \underline{272860}  &  100\%  & 281470  &  100\%  &  500000  &  64\% & -77.62\%\\
\hline g04  & 57600  &  100\%   &  \underline{30130}  &  100\%  &  36820  &  100\%  &  40140  &  100\% & -86.90\%\\
\hline g06  & 48400  &  100\%  &  \underline{11200} &  100\% &  12880  &  100\%  &  36180  &  100\% & -83.02\%\\
\hline g07  & \underline{135600}  &  100\% & 139720  &  100\%   &  135730  &  100\%  &  258840  &  100\% & -94.63\%\\
\hline g08  & 4800  &  100\%  & 2170  &  100\%  &  \underline{1960}  &  100\%  &  3510  &  100\%  & -75.41\%\\
\hline g09  & 72000  &  100\%  & 39550  &  100\%  &  \underline{37870}  &  100\%  &  58700  &  100\% & -90.92\%\\
\hline g10  & 276000  &  100\%  & 188860  &  100\%  &  325570  &  100\%  &  \underline{109970}  &  100\% & -86.60\%\\ 
\hline g12  & 15600  &  100\%  & \underline{5110}  &  100\%  &  6580  &  100\%  &  11940  &  100\% & -25.46\%\\ 
\hline g16  & 39200  &  100\%  & \underline{18970}  &  100\%  &  25060  &  100\%  &  36790  &  100\% & -83.51\%\\ 
\hline g18  & 119600  &  96\%  & 218050  &  100\%  &  134680  &  100\%  &  \underline{96180}  &  100\% & -92.44\%\\ 
\hline g19  & 212000  &  92\%  & 265930  &  100\%  &  297640  &  100\%  &  \underline{187734}  &  100\% & -86.19\%\\ 
\hline g24  & 14000  &  100\%  & \underline{5110}  &  100\%  &  5740  &  100\%  & 13690  &  100\% & -85.95\%\\ 
\hline
\end{tabular}
\renewcommand{\arraystretch}{1}
\hspace{0.3cm}
\begin{tabular}{|c|c|c|c|} \hline %m{0.71cm}
\multicolumn{4}{|c|}{\textbf{CEC 2006}} \\ 
\hline
\multirow{2}{*}{\textbf{Prob.}} & \multicolumn{2}{c|}{\textbf{Best  }} & \multirow{2}{*}{\textbf{$\Delta$NFES\%}} \\
%\multirow{3}{*}{\textbf{Prob.}} & \multicolumn{2}{c|}{\multirow{2}{*}{\textbf{BEST CEC06}}} & \multirow{3}{*}{\textbf{$\Delta$NFES\%}}\\ 
%& \multicolumn{2}{c|}{} &\\ 
\cline{2-3}
            & \textbf{NFES} & \textbf{SR} &\\ 
\hline g01  &  25115  &  100\% & -19.15\%\\ 
\hline g02  &  96222  &  100\% & -36.53\%\\ 
\hline g04  &  15281  &  100\% & -74.18\%\\ 
\hline g06  &  5202  &  100\% & -63.45\%\\ 
\hline g07  &  26578  &  100\% & -72.60\%\\ 
\hline g08  &  918  &  100\% & -47.49\%\\ 
\hline g09  &  16152  &  100\% & -78.72\%\\ 
\hline g10  &  25520  &  100\% & -42.26\%\\ 
\hline g12  &  1308  &  100\% & +191.20\%\\ 
\hline g16  &  8730  &  100\% & -64.17\%\\ 
\hline g18  &  28261  &  100\% & -74.27\%\\ 
\hline g19  &  21830  &  100\% & +18.71\%\\ 
\hline g24  &  1794  &  100\% & -59.98\%\\ 
\hline
\end{tabular}
\end{center}
}
\end{table*}
\begin{table}[!ht]
\caption{Median NFES to achieve the fixed accuracy level ($(f(\bm{x})-f(\bm{x}^*)) \leq 0.0001$) and Success Rate for the selected CEC 2006 problems (continued from Table~\ref{tab:CecComparison}). All the compared results are obtained from the corresponding papers. Results reported for GB-MA \cite{SunGaribaldi2010} show mean, rather than median NFES. $\Delta$NFES\% indicates for each problem the relative difference of NFES between \algorithmname~and the best algorithm (underlined).}\label{tab:CecComparisonNLP}
%Comparison with methods that use traditional non-linear programming techniques that access gradient information. 
{
%\tiny
\fontsize{7.2pt}{10pt}\selectfont
\setlength{\tabcolsep}{2pt}
\begin{center}
\begin{tabular}{|c|c|c|c|c|c|c|c|c|c|} \hline
\multicolumn{10}{|c|}{\textbf{MA with NLP Techniques}} \\
\hline
\multirow{2}{*}{\textbf{Prob.}} & \multicolumn{2}{c|}{\textbf{SADE \cite{Huang:SDE:cec2006} }} & \multicolumn{2}{c|}{\textbf{DMS-PSO \cite{Liang:DMP:cec2006} }} & \multicolumn{2}{c|}{\textbf{GB-MA \cite{SunGaribaldi2010} }} & \multicolumn{2}{c|}{\textbf{FSM \cite{Handoko2010} }} & \multirow{2}{*}{\textbf{$\Delta$NFES\%}} \\ 
\cline{2-9}
            & \textbf{NFES} & \textbf{SR}  & \textbf{NFES} & \textbf{SR}  & \textbf{NFES} & \textbf{SR} & \textbf{NFES} & \textbf{SR} & \\	%FSM-ac/ic
\hline g01  &  25115  &  100\%  & 25816  &  100\% &  7859  &  100\%  &  \underline{294}  &  100\%  & +6806.12\%\\ 				%294  &  100\%  \\
\hline g02  &  128970  &  84\%  & 87107  &  84\%   & \underline{45555}  &  100\% &  91324  &  100\% & +34.06\%\\				%70857  &  100\%  \\
\hline g04  &  25107  &  100\%  & 25443  &  100\%  & 1201  &  100\% &  \underline{269}  &  100\% & +1366.54\%\\ 				%224  &  100\%  \\
\hline g06  &  14404  &  100\%  & 27636  &  100\% & 489  &  100\% &  \underline{110}  &  100\% & +1628.18\%\\ 					%110  &  100\%  \\
\hline g07  &  101240  &  100\%  &  26685  &  100\%  & 3588  &  100\% &  \underline{2225}  &  100\% & +227.24\%\\ 				%3255  &  100\%  \\
\hline g08  &  1272  &  100\%  &  3892  &  100\%  & 1068  &  100\%  &  \underline{448}  &  100\% & +7.59\%\\ 					%436  &  100\%  \\
\hline g09  &  16787  &  100\%  &  29410  &  100\%  & \underline{1632}  &  100\% & 6586  &  100\% & +110.54\%\\ 				%6881 & 100\% \\
\hline g10  &  52000  &  100\%  &  25500  &  100\%  & 17319  &  100\%  &  \underline{1712}  &  100\% & +760.63\%\\				%2110  &  100\%  \\
\hline g12  &  1717  &  100\%  &  6826  &  100\%  & 348  &  100\%  &  \underline{176}  &  100\% & +2064.20\%\\ 					%118  & 100\% \text  \\
\hline g16  &  14433  &  100\%  &  28433  &  100\%  & 7092  &  100\% &  \underline{325}  &  100\% & +862.461\%\\				%325  &  100\%  \\
\hline g18  &  26000  &  92\%  &  28000  &  100\%   & 11095  &  100\% &  \underline{1875}  &  100\% & +287.84\%\\ 				%1875  &  100\%  \\
\hline g19  &  51588  &  100\%  & 21587  &  100\% & 13355 & 100\% &  \underline{1370}  &  100\% & +1791.53\%\\ 					%1152  &  100\%  \\
\hline g24  &  4843  &  100\%  &   18729  &  100\%  & 425  &  100\% &  \underline{380}  &  100\% & +88.94\%\\ 					%258  &  100\%  \\
\hline
%\multicolumn{9}{r}{(**) These algorithms use traditional non-linear programming techniques.}
\end{tabular}
\end{center}
}
\end{table}

A closer examination at the comparison between \algorithmname~and the group of algorithms derived from DE reveals a strong performance advantage of \algorithmname~in finding optimal solutions. In only one problem out of the thirteen considered, \texttt{g12}, \algorithmname~needs about 3.17 times more function evaluations to reach the optimum. In all the other problems, \algorithmname~needs between 14.1\% (\texttt{g02}) and 87.24\% (\texttt{g07}) less evaluations than the best DE-based algorithm (but, it should also be noted that on \texttt{g02} the SR of the best algorithm, namely MDE \cite{Mezura-Montes:MDE:cec2006}, is 16\%, while for \algorithmname~it is 100\%).
%0.21 (\texttt{g09}) and 0.80 (\texttt{g01}) evaluations (mean factor: 0.43 $\pm$ 0.18 SD) with respect to the other methods.

Comparing \algorithmname~with the algorithms derived from CMA-ES highlights an even stronger performance gain. \algorithmname~is slower (in terms of NFES needed to reach the optimum) only in three problems \texttt{g06}, \texttt{g08} and \texttt{g10}. It must be noted however that for \texttt{g06} and \texttt{g10} (both unimodal problems) the fastest algorithm is (1+1)-ViE, i.e. the same employed by our local search units, presented in our previous work \cite{viePPSN14}. It is therefore to be expected a somehow higher number of function evaluations, as in the method proposed here the global search component is also at work and may slow down the search for the optimum when the landscape is unimodal. On $g08$, \algorithmname~is also 1.38 times slower than APM-CMA-ES\cite{Kusakci2013a}, but given the already low NFES needed to solve this problem the performance decay in this case is not particularly relevant. In all other cases, \algorithmname~needs less evaluations (up to 95.4\% less for \texttt{g02}) than the best algorithm in the group.

Notably, \algorithmname~outperforms, on all problems, also all the PSO-based methods, as well as other algorithms not classifiable in any of the aforementioned groups (PCX\cite{Sinha:APP:cec2006}, ASRES\cite{Runarsson:AES:cec2006}, and two DE-based memetic algorithms using an adaptive policy to rank solutions depending on the population composition, namely ($\mu$+$\lambda$)-DE \cite{Wang11a} and ICDE \cite{Jia2013302}). In most cases, \algorithmname~needs around 10-20\% (or less) NFES w.r.t. the best algorithm in the two groups. This result is confirmed when \algorithmname~is compared against the best aggregate results from CEC 2006: except \texttt{g12} and \texttt{g19}, \algorithmname~is always able to solve the problem in less NFES.% except on the \texttt{g19} function where \algorithmname~is 1.2 times slower than DMS-PSO\cite{Liang:DMP:cec2006}. 

Finally, to broaden our comparative analysis we considered some modern memetic algorithms that \emph{do} make use of traditional NLP techniques, see Table \ref{tab:ComparedAlgo} (bottom) for details. These are methods that combine a population-based algorithm (DE, PSO, EDA or GA) with programming techniques, e.g. SQP. As such, these algorithms i) assume that the objective and constraint functions are differentiable and ii) compute and use first (and second) order information on constraints or objectives. Although evolutionary algorithms (such as our proposed \algorithmname) cannot be directly compared with these techniques, as they assume that no gradient information is available and are typically derivative-free, we deemed interesting to relate our method also with this part of the literature. Results of these comparisons can be found in Table \ref{tab:CecComparisonNLP}. As expected, the methods that make use of gradient information have a clear advantage in terms of NFES needed to converge to the optimum: this is particularly evident when \algorithmname~is compared against FSM \cite{Handoko2010}: the comparison shows that FSM converges in a very limited NFES, as it benefits from the use of second-order information in the SQP included in its memetic structure. On the other hand, it is remarkable that \algorithmname, despite the lack of any information about the gradient, outperforms SADE \cite{Huang:SDE:cec2006} and DMS-PSO \cite{Liang:DMP:cec2006} (which both include SQP) on all problems but two (\texttt{g12} and \texttt{g19}, respectively) and can exceed the performance of another powerful NLP-based memetic algorithm, GB-MA \cite{SunGaribaldi2010}, in four problems, namely \texttt{g08}, \texttt{g10}, \texttt{g16}, and \texttt{g18}.

\subsection{Engineering problems}

Given the particularly favorable results obtained by \algorithmname~on the CEC 2006 benchmark, we decided to test the method also on a group of well-known engineering problems. Specifically, we tested \algorithmname~on a welded beam design problem, the optimization of a pressure vessel, the design of a stepped cantilever (reference formulations for these three problems can be found in \cite{Coello2000}), and the design of a tension compression spring \cite{Gandomi11}.

\begin{table*}[!ht]
\caption{Median NFES and best fitness achieved for the engineering problems. All the compared results are 
obtained from the corresponding papers. Please note that if the median NFES to reach the optimum is not reported we show the full budget. $\Delta$NFES\% indicates the relative difference of NFES between the proposed algorithm \algorithmname~and the best algorithm that obtains the same best fitness (underlined). For the cantilever problem $\Delta$NFES\% is not defined as \algorithmname~obtains a lower fitness value than FA \cite{gandomi2011mixed}.}\label{tab:EngComparison}
{
%\tiny
\fontsize{7pt}{10pt}\selectfont
\setlength{\tabcolsep}{2pt}
\begin{center}
\begin{tabular}{|c|c|c|@{}p{2pt}@{}|c|c|c|c|c|c|c|c|c|c|c|c|c|}
\cline{1-3}\cline{5-17}
\multirow{2}{*}{\textbf{Prob.}} &\multicolumn{2}{c|}{\textbf{\algorithmname  }} & {} & \multicolumn{2}{c|}{\textbf{DETPS \cite{Zhang20131528} }} & \multicolumn{2}{c|}{\textbf{PSO-DE \cite{Liu:2010:HPS:1660170.1660493} }} & \multicolumn{2}{c|}{\textbf{MBA \cite{Sadollah20132592} }} & \multicolumn{2}{c|}{\textbf{COPSO \cite{Aguirre:2007} }} & \multicolumn{2}{c|}{\textbf{SiC-PSO \cite{Cagnina08a} }} & \multicolumn{2}{c|}{\textbf{FA \cite{gandomi2011mixed} }}  & \multirow{2}{*}{\textbf{$\Delta$NFES\%}} \\ 
\cline{2-3}\cline{5-16}
 & NFES & Fitness  & {} & NFES & Fitness  & NFES & Fitness  & NFES & Fitness  & NFES & Fitness  & NFES & Fitness  & NFES & Fitness & \\ 
\cline{1-3}\cline{5-17} 
W. Beam  &  6568  &  1.724852  & {} &  \underline{10000}  &  1.724852  &  66600  &  1.724852  &  47340  &  1.724853  &  30000  &  1.724852  &  24000  &  1.724852  &  50000  &  1.731210 &  -34.32\% \\ 
\cline{1-3}\cline{5-17}
Vessel  &  14087  &  5850.38306  & {} & 10000 &  5885.3336 &  42100  &  6059.714335  &  70650  &  5889.3216  &  30000  &  6059.714335  &  24000  &  6059.714335  &  \underline{25000}  &  5850.38306 & -43.65\%\\ 
\cline{1-3}\cline{5-17}
Spring  &  21413  &  0.012665  & {} & 10000  &  0.012665  &  42100  &  0.012665  &  \underline{7650}  &  0.012665  &  30000  &  0.012665  &  24000 &  0.012665  &  -  &  - & +179.90\%\\ 
\cline{1-3}\cline{5-17}
Cantilever  & 66894  &  63893.490839  & {} &  -  &  -  &  -  &  -  &  -  &  -  &  -  &  -  &  -  &  -  &  50000  &  63893.52 & -\\ 
\cline{1-3}\cline{5-17}
\end{tabular}
\end{center}
}
\end{table*}

For each engineering problem, we compared the median number of function evaluations needed by \algorithmname~to reach the optimum solution against the results reported in literature, as shown in Table \ref{tab:EngComparison} (again, ``-'' indicates that the result is not available). We selected algorithms from the literature considering only those reporting the values for the best solution found. Furthermore, algorithms whose performance is completely dominated by other reported ones were ignored. Notably, the number of function evaluations needed to reach on average the best solution fitness value is normally not reported in papers dealing with engineering optimization. Therefore, in those cases we report the full budget of function evaluations given to the algorithm. On two out of four problems, \algorithmname~is capable of discovering the best known solution in the lowest number of function evaluations. Interestingly, in one problem, namely the design of a cantilever beam, \algorithmname~discovered a solution which is better than the known optimum reported in the literature. On the other hand, on the spring design problem \algorithmname~converges to the known optimal solution but is almost 2.8 times slower than the fastest algorithm, MBA \cite{Sadollah20132592}. % (although we are aware that those algorithms might likely use less evaluations)

For completeness, we report in Table \ref{tab:EngineeringProblemsResults} for each problem the fitness value, the constraint violation, and the values of the design variables of the best solutions found, together with the median NFES to reach them (over 25 runs).
\begin{table}[!ht]
\caption{Fitness, constraint violation (E), and design variables ($\bm{x}^*$) 
of the best solutions discovered for the engineering problems.} \label{tab:EngineeringProblemsResults}
\renewcommand{\arraystretch}{1.1}
\begin{center}
\begin{tabular}{|p{1.2cm}|p{1.8cm}|c|p{3.35cm}|}
\hline Problem & Best Fitness \newline (Median NFES) & E & $\bm{x}^*$ \\ \hline
Welded Beam & $1.724852$ \newline  $(6568)$ & 0 &   
   $h = 0.205729627974134$ \newline
   $l = 3.470488964774360$ \newline
   $t = 9.036623829898325$ \newline
   $b = 0.205729643534243$ 
   \\
\hline
Pressure Vessel & $5850.383060$ \newline $(14087)$ & 0 &   
   $T_s = 0.75$ \newline
   $T_h = 0.375$ \newline
   $R = 38.8601036269430$ \newline
   $L = 221.3654713560083$
\\
\hline
Helical Spring & $0.012665$ \newline $(21413)$ & 0 &   
 $d =  0.051699916331388$ \newline
 $D = 0.356978944672547$ \newline
 $n = 11.273668588601133$ \\
\hline
Stepped Cantilever & $63893.490839$ \newline $(66894)$ & 0 &   
 $w_1 =   3$ \newline
 $h_1 =  60$ \newline
 $w_2 =   3.1$ \newline
 $h_2 =  55$ \newline
 $w_3 =   2.6$ \newline
 $h_3 =  50$  \newline
 $w_4 = 2.204553242032800$
 $h_4 =  44.091064840656017$
 $w_5 = 1.749768676906988$
 $h_5 =   34.995373538139674$\\  
\hline
\end{tabular}
\end{center}
\end{table}

% The mapping for the stepped cantilever is 
% x1,2 = b4 h4  x3,4 = b5, h5  x5,6 = b1, h1   x7,8 = b2,h2 ...

\subsection{Sample algorithm runs}

\begin{figure*}[!t]
\centering
\includegraphics[width=16cm]{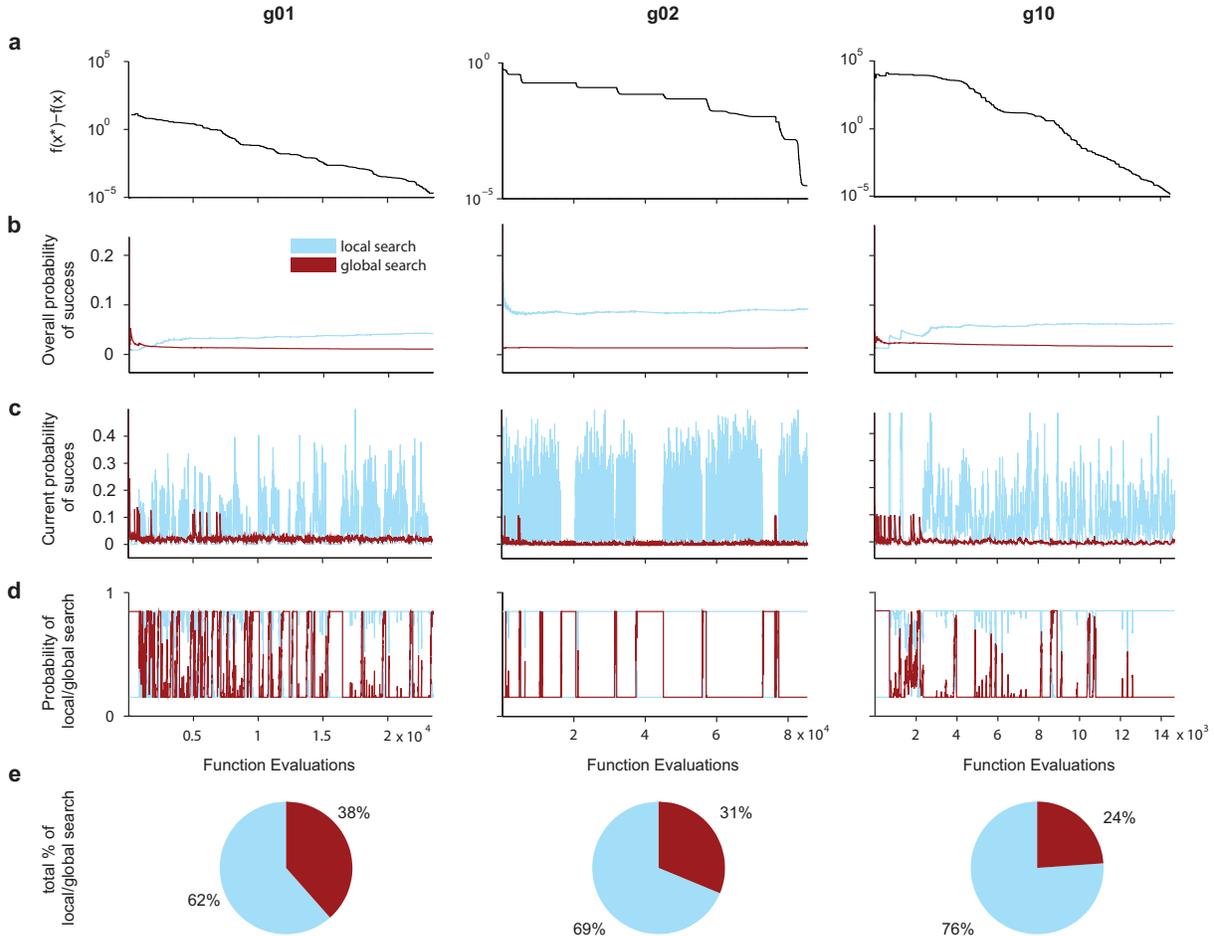}
\caption{Sample execution of the method on three functions taken from the CEC 2006 benchmark, namely \texttt{g01}, \texttt{g02}, and \texttt{g10}. We show \textbf{a}) the distance from the optimum, \textbf{b}) the overall probability of success of local and global search, \textbf{c}) the moving average of the probability of success of the two components, and \textbf{d}) their resulting frequency of execution. Finally, panel \textbf{e}) shows the resulting allocation of function evaluations to local and global search.}
\label{fig:PerformanceDissection}
\end{figure*}

Overall, \algorithmname~compared very favorably on both the CEC 2006 benchmark problems and the four selected engineering problems. To provide an idea of the algorithm dynamics, we show in Fig. \ref{fig:PerformanceDissection} the sample execution of our method on three selected problems, respectively \texttt{g01}, \texttt{g02} and \texttt{g10}, where the latter one is unimodal. The convergence plots (Fig. \ref{fig:PerformanceDissection}a) and the variables used by the scheduler (Fig. \ref{fig:PerformanceDissection}b-c) to make informed decisions on the frequency of selection of local versus global search are shown. We also display the actual probability of executing local/global search (Fig. \ref{fig:PerformanceDissection}d) and the total number of function evaluations performed by local and global search (Fig. \ref{fig:PerformanceDissection}e). It is noticeable how on the unimodal problem (\texttt{g10}) a higher number of function evaluations are allocated to the local search units. Furthermore, problems characterized by the presence of distinct local optima (\texttt{g02}) present periods in which \algorithmname~tries to locally optimize a local optimum intertwined with phases of global exploration during which the algorithm is capable of escaping from local optima. In all the three cases, we should note that the limits imposed on the frequency of selection of the global/local steps are necessary to avoid the algorithm falling in a phase where only global or local search is used, without the possibility to switch.

\subsection{Performance dissection}

\begin{figure}[!h]
\centering
\includegraphics{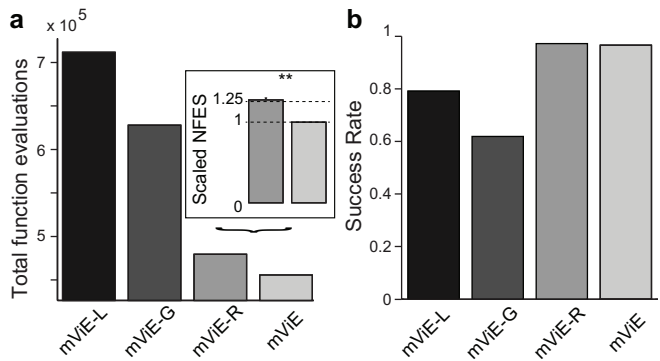}
\caption{Performance dissection over 25 runs of each of the selected CEC 2006 problems for the four experimental conditions. {\bf a}) Total number of function evaluations (aggregate over all runs and problems) to reach for each problem the target fitness difference from the optimum ($10^{-4}$). If the target is not reached, the maximum budget (250000 NFES) is added. The inset shows the scaled NFES, calculated as mean (over all problems) of the median NFES (calculated independently for each problem, considering only  successful runs), scaled by the corresponding median NFES of \algorithmname. \algorithmname~results $\sim25\%$ more efficient than \algorithmname -R (Wilcoxon Rank-Sum test, $\alpha=0.05, p=0.0015$). {\bf b}) Average success rate over all runs and problems.}
\label{fig:PerformanceComponents}
\end{figure}

Finally, we conclude this section by showing the contribution of the algorithmic components (local search, global search and scheduler) to the overall performance of \algorithmname. To assess this, we performed three separated experiments. First, in the condition \algorithmname -L we enabled only the local search component. Multiple units (with restart) were still used, but without applying global search. Second, in the condition \algorithmname -G no local search unit was stepped, and only the global search component (DE operators) was used to explore the search space. Finally, in the third condition \algorithmname -R both local and global components were used, coordinated by a trivial scheduler that randomly chooses the execution of global or local search at each iteration. We executed the algorithms corresponding to the three conditions for 25 runs on the selected CEC 2006 benchmark problems. Results are reported in Fig. \ref{fig:PerformanceComponents}, that shows the comparison of the three conditions against the fully-featured \algorithmname. 

A rather evident result is that the success rates of the local and global search component alone are lower than those of the algorithms using the two components together, as expected. Furthermore, the amount of function evaluations used to discover the optimum is very high in the \algorithmname -L condition due to the high number of restarts used by the algorithm before discovering a successful solution. Remarkably, the use of the two components allows the discovery of the optimum most of the time, even using a random scheduler (\algorithmname -R). However, the introduction of the adaptive scheduler allows a further $\sim25\%$ reduction in the number of function evaluations used.

\section{Discussion and Conclusions}\label{sec:conclusion}

In this paper we presented \algorithmname, a novel memetic computing approach for constrained optimization based on Viability Evolution. 
The proposed method is composed of multiple (1+1)-CMA-ES acting as local optimizers, that are combined with a Differential Evolution scheme to perform global search. 

In numerical experiments, our method displayed a particularly robust performance, solving a broad number of test functions (Tables~\ref{tab:CecInequalitySummary}, \ref{tab:CecComparison}) and engineering design problems (Tables~\ref{tab:EngComparison}, \ref{tab:EngineeringProblemsResults}) more efficiently than the state-of-the-art compared methods, both in terms of success rate and number of function evaluations needed to reach the optimum. %Excluding the gradient-based memetic algorithm proposed in \cite{SunGaribaldi2010}, whose performance obviously benefits from the use of gradient information, 
\algorithmname~outperformed the DE-based methods in all the comparisons (except for one case, \texttt{g12}), as well as the CMA-ES-based algorithms (except for three cases, \texttt{g06}, \texttt{g08} and \texttt{g10}), the variants of PSO (except in one case, \texttt{g19}) and all the other algorithms under consideration.

Interestingly, the few cases in which \algorithmname~was outperformed correspond to problems where either strongly explorative algorithms (favoring global search only) or strongly exploitative ones (favoring local search) perform well enough. For example, the comparison with the algorithm we proposed in our previous work \cite{viePPSN14} revealed that on fitness landscapes such as \texttt{g06} and \texttt{g10} a single local search unit performs more efficiently than multiple ones coordinated by DE (as in \algorithmname). We believe that in these cases the additional overhead introduced by the initial learning phase is the cause of this performance decay. On the other hand, under the general assumption that there is no prior knowledge on the features of the landscape, a learning phase might be necessary. Also, most of the problems are actually characterized by fitness landscapes where a proper trade-off between global and local search has to be found, and these are the cases where \algorithmname~excel.

In our opinion, the reason for the success of \algorithmname~is twofold: first, in the local search units the modelling of constraints as viability boundaries allows the search to be driven towards the feasible space and its most promising areas; secondly, the adaptive scheduler coordinates both the local search units and global recombination, thus enabling a synergistic exploration of the constrained search space. We discuss in detail these two aspects below. 

\subsubsection{Modelling of constraints as viability boundaries} as the local search units define viability boundaries separately on every constraint, they can collect additional information on each constraint. This information is therefore beneficial for a more fruitful and faster search, for example by adapting step-size and covariance matrix. On the contrary, approaches that combine constraints into the objective function or as a single constraint violation measure lose a potential source of information.

Dealing with constrained optimization problems without aggregating the constraints in the fitness function or in an aggregate constraint violation function unlocks additional information that is readily available to an evolutionary algorithm. Still, the large majority of approaches in the literature make use of some form of aggregation of constraints\footnote{We also aggregate the constraints into a single constraint violation function when applying global search operators and when comparing against the global best solution in the main algorithmic loop. In a future version of \algorithmname~it may be beneficial to remove as well this form of aggregation.}.
We therefore deemed useful to ask the question whether or not the current abstraction under which evolutionary algorithms operate is the most appropriate for optimization problems different from unconstrained ones. The logic behind most of the available constraint handling techniques, such as penalty functions, is in our view symptomatic of how the current evolutionary algorithms paradigm may be misleading in the design of novel algorithms. Under this traditional paradigm, algorithms are designed for having solutions in competition with each other based on their fitness function values. Solutions are therefore ranked and compared uniquely using this single value. Intuitively modelled on a very high-level abstraction of natural evolution, this paradigm may hinder the development of the field towards more comprehensive paradigms and thus more powerful algorithms. In fact, the current abstraction forces researchers and adopters into thinking an evolutionary process as naturally modelled using a single fitness function.

%An alternative paradigm of artificial evolution, called %\cite{Mattiussi2003,maesani2014artificial} 
On the other hand, Viability Evolution models an evolutionary process as elimination of solutions that do not comply with certain viability criteria, defined on both problem objectives and constraints. By adapting these criteria during the search, it is possible to drive the solutions towards desired areas of search space, typically the global optimum or feasible regions.

A first direct implication of the Viability Evolution paradigm is that constraints and objectives are kept implicitly separated\footnote{%In a similar sense, another shift in paradigm, the design of multi-objective methods and the introduction of Pareto optimality in constrained optimization, helped the field in developing powerful algorithms.
A similar shift in paradigm was observed in multi-objective evolutionary algorithms (MOEAs), when classic aggregation methods such as weighted summing of objectives were replaced by the use of Pareto optimality concepts. Such a shift led to radically novel MOEAs, eventually obtaining dramatic performance improvements.}. Also, the ``fitness'' of individuals is in this case a property measured \textit{a-posteriori} and not defined \textit{a-priori} as done by using a fitness function in the classic sense. Furthermore, the availability of statistics on the viability of solutions made possible by this different abstraction, e.g. the number of individuals satisfying specific viability criteria or the number of viable/non-viable individuals, leads to increased information available for evolutionary methods. 
Third, more emphasis on elimination of non-viable solutions rather than competition of solutions by a unique fitness score may lead to enhanced diversity in the evolving population \cite{maesani2014artificial}.

Here, we showed that by following the design principles of Viability Evolution it is possible to derive a very efficient method for constrained optimization. Overall our method models constraints separately and more importantly uses information about constraint violations (non-viability of individuals) for adapting the algorithm parameters during the search, without requiring a user to aggregate them.

\subsubsection{Global recombination and memetic adaptation} a key element of \algorithmname~is the scheduler, that adaptively activates local or global search, providing a proper balance between the two regimes. An important advantage of this scheme is the fact that it has \emph{direct} control over the budget assigned to both the global and local search operators. This feature makes our scheduler different from most of the existing credit assignment mechanisms used in memetic algorithms, which in general balance the budget assigned to different local search methods (memes), while they do not regulate directly the budget assigned to the global search operators, see e.g. \cite{ong2004metalamlearn}. One exception is the method presented in \cite{nguyen2009}, which uses a measure called \emph{local search intensity} for allocating the budget assigned to each meme in a pool of local search methods. However, also in this case the control over the global search budget is implicit rather than explicit, i.e. global and local search are always executed at each step and only the number of function evaluations allotted to local search is adjusted. We believe, instead, that a direct control of both global and local search function evaluation budgets is of fundamental importance especially in constrained optimization scenarios, where it may be needed to adjust dynamically the exploration pressure. Overall, our algorithm provides two levels of adaptation: the first level is represented by our scheduler, whereas the second level consists of the use of self-adapting local search units, that learn their parameters during the search. 
%Conversely, in memetic computing methods, the parameters of the memes are fixed and do not adapt during the search. 
Following the taxonomy provided in \cite{ong2006}, our method can be therefore classified as a \emph{self-adaptive} algorithm (because multiple (1+1)-ViE units coevolve, adapting independently their parameters), with adaptation at both \emph{local} and \emph{global} level: our method employs in fact a mix of recent information on the performance of local and global search (the moving averages of the probabilities of success) and complete historical information (the absolute number of successes of each operator).
\newline

To conclude, the present work contributes to the field of constrained optimization and suggests a wide spectrum of 
possible research lines that are worth following, going in the direction of: 1) extending the viability concept to different 
classes of problems, such as large-scale, multi-objective, and dynamic optimization; 2) testing alternative recombination schemes, 
based for instance of swarm intelligence, to coordinate the multiple local search operators; and 3) applying the proposed method to real-world applications where a resource-efficient constrained optimization solver might be needed, for instance in various domains of engineering or computational biology.

% or


% If adding more appendices, use sections and decomment appendices
\appendices
\section{Parameter Analysis}\label{apx:parameter-study}

Given the impossibility of performing a full combinatorial exploration of the parameter space, we executed a preliminary tuning of the main parameters by following an iterative procedure. First, we set the initial parameters by empirically experimenting with the algorithm. We then analyzed the influence on the algorithmic performance varying each parameter, fixing the identified optimal parameter values in several steps. We investigated independently $c_\alpha$ and $\beta_R$, secondly $L$, $F$ and $CR$, and finally $pop_{size}$.

We evaluated the algorithm's performance by testing it on the full CEC 2006 benchmark (with inequalities only) for 25 runs. We allowed the algorithm to run for 150000 function evaluations. We measured the success rate (SR) for each function and the factor of number of function evaluations (NFES) with respect to the ones achieved by the best algorithm of the CEC 2006 competition to reach the optimum at the accuracy of $10^{-4}$. To obtain a more robust evaluation of the success rate, we tested each parameter combination 5 times, for a total of 5 repetitions $\times$ 25 runs $ \times$ 13 benchmark problems. We aggregated SR and NFES factor across all problems: we summed the rank of a parameter configuration calculated considering the success rate (higher SR are assigned better ranks) and the NFES factor (lower NFES factors are assigned better ranks). To compute a single score for each parameter configuration, we summed the aggregate rank obtained by each parameter setting on all the tested problems.

\begin{figure*}[!pht]
\centering
\includegraphics[width=16cm]{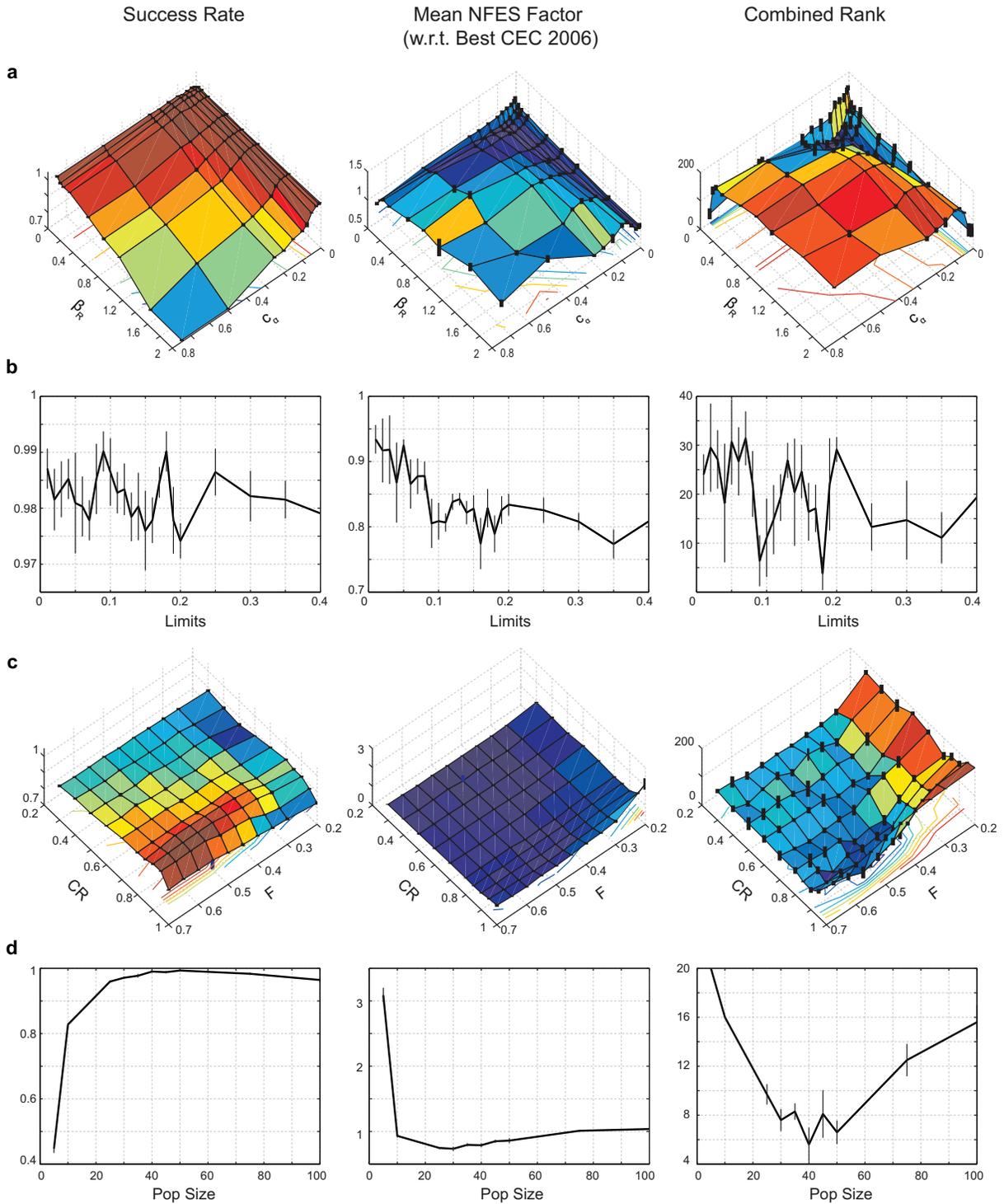}
\caption{Parameter analysis: each reported result is the aggregation obtained by repeating 25 times each function (with inequalities only) of the CEC 2006 benchmark for 5 independent repetitions (i.e. 13 benchmark functions $\times$ 25 runs $\times$ 5 repetitions). The standard deviation across the 5 repetitions is reported as a black bar for each parameter combination. We show mean success rate across the 13 benchmark functions and mean factor of function evaluations computed with respect to the best algorithm in the CEC 2006 competition. To select the best parameter combination for each of the parameters, we independently rank each parameter combination by success rate and mean factor of function evaluations. In the third column, we show the combined (summation) rank for each parameter combination. We report four parameter studies for {\bf a}) parameters regulating the update rules of the success probability, $c_\alpha$ and $\beta_R$; {\bf b}) minimum global/local search execution frequency limit $L$; {\bf c}) DE operators' parameters $F$ and $CR$; and {\bf d}) number of local search units, i.e. population size $pop_{size}$.}
\label{fig:ParameterStudy}
\end{figure*}

The results of this parameter tuning procedure are shown in Fig. \ref{fig:ParameterStudy}. Success rate, mean factor of NFES and aggregate rank of each parameter combination are shown when varying $c_\alpha$ and $\beta_R$ (Fig. \ref{fig:ParameterStudy}a), the minimum relative execution frequency limit $L$ (Fig. \ref{fig:ParameterStudy}b), $F$ and $CR$ (Fig. \ref{fig:ParameterStudy}c), and the population size (Fig. \ref{fig:ParameterStudy}d). The final identified parameter values are $c_\alpha = 0.1, \beta_R = 0.05, L = 0.18, F = 0.5, CR = 0.9, P_{size} = 40$. Although this may be a locally optimal combination of parameters and better parameter tuning may be achieved by a more thorough parameter exploration, we considered this procedure sufficient for our purposes. Also, we obtained an idea of how the method's performance is affected by changing its parameters.

More specifically, the first two parameters ($c_\alpha$ and $\beta_R$), which regulate the way the success probability of global and local search is updated (which, in turn, affect the information used by the scheduler to allocate function evaluations to the two components), seem to have a large impact on the performance. We can see from Fig. \ref{fig:ParameterStudy}a that the highest values of SR are obtained when at least one of the two parameters has a very low value, namely $c_\alpha \in [0,0.1]$ or $\beta_R \in [0,0.1]$. Values lying at the intersection of these two regions, e.g. $c_\alpha=0.1$ and $\beta_R=0.05$ (the values we chose in our experimentation), provide then among the highest SR. Interestingly, these values provide also the lowest NFES factor, and therefore the best rank. We further verified this setting looking at the performance on each single problem obtained with different values of $c_\alpha$ and $\beta_R$ (plots are not reported here for brevity). Interestingly, the SR landscape has a completely flat surface for \texttt{g1}, \texttt{g4}, \texttt{g6}, \texttt{g8}, \texttt{g12}, \texttt{g16} and \texttt{g24}. In the other cases, it has a plateau with maximum values on the borders and a pronounced valley towards the region $c_\alpha>0.3$ and $\beta_R>1$. For almost all problems (apart from \texttt{g19}), the NFES factor landscape has a large plateau with maxima (for \texttt{g1}, \texttt{g2} and \texttt{g12}) or minima (all other cases) at the border regions. Overall, the setting corresponding to values in $[0.05,0.1]$ for both $c_\alpha$ and $\beta_R$ is considered the most robust and should be used in practical applications. Higher values for both parameters make the previous information on the success probability fade faster (see Equations \ref{eq:increaseProb}-\ref{eq:decreaseProbGlob}), thus reducing the effect of the learning for the scheduler and producing a less efficient switch between global and local search.

As seen in Fig. \ref{fig:ParameterStudy}b, the limit $L$ on the frequency of activation of global/local search has a very limited impact on the performance. For $L \in [0,0.4]$, the success rate is always between 0.98 and 0.99, while the NFES factor seems to be a bit lower for $L \ge 0.15$. The combined ranking information does not show a clear pattern in this case. We chose the value $L=0.18$ because it seems to provide the best trade-off SR-NFES. In general, we recommend using the setting $L \in [0.15,0.4]$, which should be relatively robust on different classes of problems. %Again, we verified this by looking at the performance on each single problem obtained with different values of L (not reported here for brevity).
Values outside this range make the frequency of activation of either local or global search too low ($L<0.15$) or too high ($L>0.4$) thus unbalancing the NFES allocated to the two search regimes.

Considering $F$ and $CR$, i.e. the two parameters involved in the DE-based global search operators, we know from the literature \cite{book:price2005} that a generally recommended setting is $F=0.5$ and $CR=0.9$. Nevertheless, we verified this setting experimentally as seen in Fig. \ref{fig:ParameterStudy}c. From the figure, we can observe that the SR landscape is characterized by a peak in the region $F \in [0.4,0.7]$ and $CR \in [0.8,1.0]$. The NFES factor landscape has instead a large plateau with a small peak corresponding to $F \in [0.2,0.3]$ and $CR \in [0.9,1.0]$. Finally, the rank landscape (aggregating the info on SR and NFES) suggests that the minimum rank (i.e., the best rank) is in the region around $CR=0.9$ and $F \in [0.5,0.6]$. Overall, our experiments confirm that the most robust setting corresponds to $F=0.5$ and $CR=0.9$. We believe that this setting should be used in all applications of our method. Using different values for either $F$ or $CR$ has the effect of making mutation and crossover too disruptive (or not effective enough).

Finally, as for the population size (i.e., the number of local search units), we can see from Fig. \ref{fig:ParameterStudy}d that the parameter value chosen, $pop_{size}=40$, guarantees the best trade-off SR-NFES for all problems. Indeed, the SR hits the maximum at $pop_{size}=40$, while the NFES factor has a minimum for $pop_{size} \in [30,40]$. Overall, we suggest using values in this range, which seem robust enough to handle all the problems. Intuitively, smaller populations have a more limited exploration and may fail on highly multi-modal landscapes. Also too large populations have in our case a limited advantage, as the initial probability of selecting one local search unit decreases with the number of units in the population, and in later stages it is likely that only few successful units are selected.
\section{CEC 2006 problem results - Error values achieved at different level of NFES}\label{apx:error-values}

Table \ref{tab:CecErrorSummary} reports the best, median, worst and mean error values, i.e. the difference between the known optimum fitness and the best fitness achieved by \algorithmname, at different NFES. Furthermore, the number of violated constraints $c$ and the average sum of constraint violation $\bar{v}$ at the median solution (at the end of each NFES level) are shown.

\begin{table*}[!ht]
\caption{Error values (difference between the known optimum fitness and the best fitness value) achieved when NFES= $5\times10^3$, NFES= $5\times 10^4$, and NFES= $5\times 10^5$ for the selected CEC 2006 problems. Best, median, worst and mean error values are reported in the table, together with the number of violated constraints $c$ and the average sum of constraint violation $\bar{v}$ of the median solution.
} \label{tab:CecErrorSummary}
\renewcommand{\arraystretch}{1.1}
\scriptsize
\begin{center}
\begin{tabular}{|c|c|c|c|c|c|c|}\cline{1-7} 
%& & & & & & \\
{\bf NFES}&   & {$\mathrm g01$}& {$\mathrm g02$}& {$\mathrm g04$}& {$\mathrm g06$}& {$\mathrm g07$}\\
%& & & & & & \\
\cline{1-7}                         & Best& 0.695563 (0)& 0.132752 (0)& 0 (0)& 0 (0)& 0.001159 (0)\\
\cline{2-7}                         & Median& 1.91028 (0)& 0.302738 (0)& 1e-06 (0)& 0 (0)& 0.015729 (0)\\
\cline{2-7}                         & Worst& 3.7779 (0)& 0.418978 (0)& 0.202275 (0)& 0.02595 (0)& 1.47459 (0)\\
\cline{2-7} $\mathbf{5\times 10^3}$ & $c$& 0, 0, 0& 0, 0, 0& 0, 0, 0& 0, 0, 0& 0, 0, 0\\
\cline{2-7}                         & $\overline{v}$& 2.429400e-02& 0& 0& 6.130000e-04& 2.052570e-01\\
\cline{2-7}                         & Mean& 2.1593& 0.287957& 0.00825104& 0.00141272& 0.110063\\
\cline{2-7}                         & Std& 0.822658& 0.0663392& 0.0404259& 0.00529882& 0.307058\\
\cline{1-7}                         & Best& 0 (0)& 0 (0)& 0 (0)& 0 (0)& 0 (0)\\
\cline{2-7}                         & Median& 0 (0)& 0.011011 (0)& 0 (0)& 0 (0)& 0 (0)\\
\cline{2-7}                         & Worst& 7.6e-05 (0)& 0.132974 (0)& 0 (0)& 0 (0)& 0.0001 (0)\\
\cline{2-7} $\mathbf{5\times 10^4}$ & $c$& 0, 0, 0& 0, 0, 0& 0, 0, 0& 0, 0, 0& 0, 0, 0\\
\cline{2-7}                         & $\overline{v}$& 4.013410e-01& 0& 0& 1.375290e-01& 3.400200e-01\\
\cline{2-7}                         & Mean& 1.968e-05& 0.0228126& 0& 0& 4e-06\\
\cline{2-7}                         & Std& 2.7308e-05& 0.0296265& 0& 0& 2e-05\\
\cline{1-7}                         & Best& 0 (0)& 0 (0)& 0 (0)& 0 (0)& 0 (0)\\
\cline{2-7}                         & Median& 0 (0)& 0 (0)& 0 (0)& 0 (0)& 0 (0)\\
\cline{2-7}                         & Worst& 0 (0)& 0 (0)& 0 (0)& 0 (0)& 0 (0)\\
\cline{2-7} $\mathbf{5\times 10^5}$ & $c$& 0, 0, 0& 0, 0, 0& 0, 0, 0& 0, 0, 0& 0, 0, 0\\
\cline{2-7}                         & $\overline{v}$& 4.964673e+00& 0& 6.900000e-05& 8.971150e+00& 4.285058e+00\\
\cline{2-7}                         & Mean& 0& 0& 0& 0& 0\\
\cline{2-7}                         & Std& 0& 0& 0& 0& 0\\
\cline{1-7} 
\multicolumn{7}{c}{} \\
\cline{1-7} 
%& & & & & & \\
{\bf NFES}&   & {$\mathrm g08$}& {$\mathrm g09$}& {$\mathrm g10$}& {$\mathrm g12$}& {$\mathrm g16$}\\
%& & & & & & \\
\cline{1-7}                         & Best& 0 (0)& 0 (0)& 10.0031 (0)& 0 (0)& 0 (0)\\
\cline{2-7}                         & Median& 0 (0)& 0 (0)& 216.449 (0)& 0 (0)& 0 (0)\\
\cline{2-7}                         & Worst& 0 (0)& 0.410366 (0)& 4434.49 (0)& 0.005625 (0)& 0.000864 (0)\\
\cline{2-7} $\mathbf{5\times 10^3}$ & $c$& 0, 0, 0& 0, 0, 0& 0, 0, 0& 0, 0, 0& 0, 0, 0\\
\cline{2-7}                         & $\overline{v}$& 9.640000e-04& 0& 1.444769e+03& 0& 1.902000e-03\\
\cline{2-7}                         & Mean& 0& 0.0224805& 922.665& 0.00116472& 8.572e-05\\
\cline{2-7}                         & Std& 0& 0.0833781& 1447.02& 0.00227881& 0.000239219\\
\cline{1-7}                         & Best& 0 (0)& 0 (0)& 0 (0)& 0 (0)& 0 (0)\\
\cline{2-7}                         & Median& 0 (0)& 0 (0)& 3e-06 (0)& 0 (0)& 0 (0)\\
\cline{2-7}                         & Worst& 0 (0)& 0 (0)& 0.265618 (0)& 0 (0)& 0 (0)\\
\cline{2-7} $\mathbf{5\times 10^4}$ & $c$& 0, 0, 0& 0, 0, 0& 0, 0, 0& 0, 0, 0& 0, 0, 0\\
\cline{2-7}                         & $\overline{v}$& 1.057000e-03& 0& 2.776000e-03& 2.800000e-05& 1.864644e+01\\
\cline{2-7}                         & Mean& 0& 0& 0.01235& 0& 0\\
\cline{2-7}                         & Std& 0& 0& 0.0530731& 0& 0\\
\cline{1-7}                         & Best& 0 (0)& 0 (0)& 0 (0)& 0 (0)& 0 (0)\\
\cline{2-7}                         & Median& 0 (0)& 0 (0)& 0 (0)& 0 (0)& 0 (0)\\
\cline{2-7}                         & Worst& 0 (0)& 0 (0)& 0 (0)& 0 (0)& 0 (0)\\
\cline{2-7} $\mathbf{5\times 10^5}$ & $c$& 0, 0, 0& 0, 0, 0& 0, 0, 0& 0, 0, 0& 0, 0, 0\\
\cline{2-7}                         & $\overline{v}$& 1.057000e-03& 9.073334e+00& 3.134842e+03& 2.800000e-05& 1.082927e+02\\
\cline{2-7}                         & Mean& 0& 0& 0& 0& 0\\
\cline{2-7}                         & Std& 0& 0& 0& 0& 0\\
\cline{1-7} 
\multicolumn{7}{c}{} \\
\cline{1-5} 
%& & & & \\
{\bf NFES}&   & {$\mathrm g18$}& {$\mathrm g19$}& {$\mathrm g24$}\\
%& & & & \\
\cline{1-5}                         & Best& 5e-05 (0)& 13.4913 (0)& 0 (0)\\
\cline{2-5}                         & Median& 0.010392 (0)& 43.6379 (0)& 0 (0)\\
\cline{2-5}                         & Worst& 0.595968 (3)& 107.837 (0)& 0 (0)\\
\cline{2-5} $\mathbf{5\times 10^3}$ & $c$& 0, 0, 0& 0, 0, 0& 0, 0, 0\\
\cline{2-5}                         & $\overline{v}$& 4.908650e-01& 0& 0\\
\cline{2-5}                         & Mean& 0.095619& 46.6881& 0\\
\cline{2-5}                         & Std& 0.154321& 19.8156& 0\\
\cline{1-5}                         & Best& 0 (0)& 0 (0)& 0 (0)\\
\cline{2-5}                         & Median& 0 (0)& 0 (0)& 0 (0)\\
\cline{2-5}                         & Worst& 0.191044 (0)& 2e-06 (0)& 0 (0)\\
\cline{2-5} $\mathbf{5\times 10^4}$ & $c$& 0, 0, 0& 0, 0, 0& 0, 0, 0\\
\cline{2-5}                         & $\overline{v}$& 4.561900e-01& 0& 1.494000e-03\\
\cline{2-5}                         & Mean& 0.0152835& 8e-08& 0\\
\cline{2-5}                         & Std& 0.0528977& 4e-07& 0\\
\cline{1-5}                         & Best& 0 (0)& 0 (0)& 0 (0)\\
\cline{2-5}                         & Median& 0 (0)& 0 (0)& 0 (0)\\
\cline{2-5}                         & Worst& 0 (0)& 0 (0)& 0 (0)\\
\cline{2-5} $\mathbf{5\times 10^5}$ & $c$& 0, 0, 0& 0, 0, 0& 0, 0, 0\\
\cline{2-5}                         & $\overline{v}$& 7.948723e+00& 0& 4.546000e-03\\
\cline{2-5}                         & Mean& 0& 0& 0\\
\cline{2-5}                         & Std& 0& 0& 0\\
\cline{1-5} 
\multicolumn{7}{c}{} \\
\end{tabular}
\end{center}
\end{table*}

%\vspace{1cm}

\section*{Acknowledgment}
We thank Pavan Ramdya for useful comments on the manuscript. The computations were performed on the EPFL HPC Cluster 
``Aries'' (\url{http://scitas.epfl.ch}). This research has been supported by the Swiss National Science Foundation, grant number 141063. 
%127143 (viability) 141063 (dispersal and altruism)

%\clearpage
%\vfill\break
%\ifCLASSOPTIONcaptionsoff
%  \newpage
%\fi

%\clearpage\clearpage\newpage





%\vspace{1cm}
\IEEEtriggeratref{108}
\bibliographystyle{IEEEtran}
\bibliography{references}

\end{document}